\documentclass[]{fairmeta}

\usepackage[utf8]{inputenc}

\usepackage{amsmath,amsfonts,bm}

\def\eqref#1{equation~\ref{#1}}

\def\1{\bm{1}}

\DeclareMathAlphabet{\mathsfit}{\encodingdefault}{\sfdefault}{m}{sl}
\SetMathAlphabet{\mathsfit}{bold}{\encodingdefault}{\sfdefault}{bx}{n}

\usepackage{wrapfig}
\usepackage{tabularx}
\usepackage{textcomp}
\usepackage{stfloats}
\usepackage{url}
\usepackage{verbatim}
\usepackage{titlesec}
\usepackage{tocloft}
\usepackage{adjustbox}
\usepackage{multirow}
\usepackage{pifont}
\usepackage{tikz}
\usepackage{comment}
\usepackage{amsmath,amssymb,amsfonts}
\usepackage{colortbl}
\usepackage{color}
\usepackage{booktabs}
\usepackage{graphicx}
\usepackage{subcaption}
\usepackage{array}
\usepackage{makecell}
\usepackage{overpic}
\usepackage{threeparttable}
\usepackage{float}
\usepackage{xspace}
\usepackage{listings}

\preto\section{\FloatBarrier}

\definecolor{headerbg}{RGB}{240,244,248}
\definecolor{groupbg}{RGB}{240,244,248}
\definecolor{oursbg}{RGB}{234,247,236}
\definecolor{softgreen}{RGB}{234,247,236}
\definecolor{bestbg}{RGB}{235,243,252}
\definecolor{softblue}{RGB}{235,243,252}
\definecolor{lossbg}{RGB}{252,235,235}
\definecolor{softred}{RGB}{252,235,235}
\definecolor{goodgreen}{RGB}{34,139,34}
\definecolor{badred}{RGB}{192,57,43}
\definecolor{softgray}{RGB}{100,100,100}

\newcommand{\gup}[1]{\textcolor{goodgreen}{$\uparrow\,#1$}}
\newcommand{\gdown}[1]{\textcolor{goodgreen}{$\downarrow\,#1$}}
\newcommand{\rup}[1]{\textcolor{badred}{$\uparrow\,#1$}}
\newcommand{\rdown}[1]{\textcolor{badred}{$\downarrow\,#1$}}

\newcommand{\gain}[1]{\gup{#1}}

\newcommand{\answerTODO}[1][]{\textcolor{red}{\bf [TODO]}}
\newcommand{\justificationTODO}[1][]{\textcolor{red}{\bf [TODO]}}

\renewcommand{\paragraph}[1]{\vspace{1.25mm}\noindent\textbf{#1}}

\setlabdisplayname{OmniAI Group of ZJU ACES Lab}
\setuniversityname{Zhejiang University}

\title{VLA-Corrector: Lightweight Detect-and-Correct Inference for Adaptive Action Horizon}

\author[1]{Yi Pan}
\author[1]{Miao Pan}
\author[1]{Qi Lu}
\author[1]{Jiaming Huang}
\author[1]{Man Zhang}
\author[2]{Siteng Huang}
\author[2]{Xin Li}
\author[1]{Jie Zhang}
\author[1]{Yongliang Shen}
\author[1]{Xuhong Zhang}
\author[1]{Wenqi Zhang}

\affiliation[1]{Zhejiang University}
\affiliation[2]{Alibaba DAMO Academy}
\metadata[Yi Pan]{\email{panyi0304@gmail.com}}
\metadata[Wenqi Zhang]{\email{zhangwenqi@zju.edu.cn}}

\abstract{
Vision-Language-Action (VLA) foundation models have recently achieved strong progress in embodied intelligence. To reduce policy-call frequency while preserving temporal coherence, most generative policies adopt an \textbf{action chunk} mechanism, executing multiple future actions in an open-loop manner under a fixed \textbf{action horizon}. However, this ``predict-then-blindly-execute'' paradigm sacrifices closed-loop reactivity: in contact-rich physical interactions, even small local perturbations can rapidly amplify within the open-loop blind spot, leading to compounding errors and ultimately task failure.
To address this limitation, we propose \textbf{VLA-Corrector}, a lightweight corrective inference framework for action-chunked VLA policies. Without modifying the backbone policy weights, VLA-Corrector introduces a lightweight \textbf{Latent-space Vision Monitor} (LVM) that continuously compares predicted and actual visual feature evolution, enabling online detection of visual dynamics deviations. Once persistent deviation is detected, the system triggers a truncation event, discards the remaining stale actions, and invokes corrective replanning via \textbf{Online Gradient Guidance} (OGG).
The detect-and-correct mechanism of VLA-Corrector naturally induces an \textbf{event-triggered adaptive action horizon}: it preserves long-horizon execution when the current chunk remains reliable, and invokes short-horizon corrective replanning when execution begins to drift. In doing so, VLA-Corrector mitigates the trade-off imposed by static horizons between execution robustness and policy-call frequency. It can be integrated into different VLA models without further retraining the VLA backbone, interrupting compounding errors while preserving much of the efficiency benefit of action chunking and substantially improving robustness in long-horizon, contact-rich robotic manipulation tasks.
Our code is available at \url{https://github.com/ZJU-OmniAI/vla-corrector}.
}

\date{October 2026}

\begin{document}
\thispagestyle{firstheader}
\maketitle
\pagestyle{plain}

\clearpage
\tableofcontents
\clearpage

\section{Introduction}

Vision-Language-Action (VLA) foundation models have emerged as a promising path toward general-purpose robot control, unifying perception, language, and action generation within a single framework~\cite{zitkovich2023rt,o2024open,kim2024openvla,black2024pi_0,li2023vision,bjorck2025gr00t,intelligence2025pi_,ma2024survey,wang2026vla,cen2025rynnvla,jiang2025rynnvla}. 
Modern VLA policies often rely on generative action models, such as diffusion models and flow matching, to capture the high-dimensional and multi-modal nature of continuous robot actions~\cite{black2024pi_0,chi2025diffusion,jiang2025streaming,wang2024one,lipman2022flow}. 
However, the per-step latency of generative models precludes closed-loop replanning, creating a fundamental tension between action expressiveness and high-frequency feedback control.

\begin{figure}[t]
    \centering
    \includegraphics[width=0.95\columnwidth]{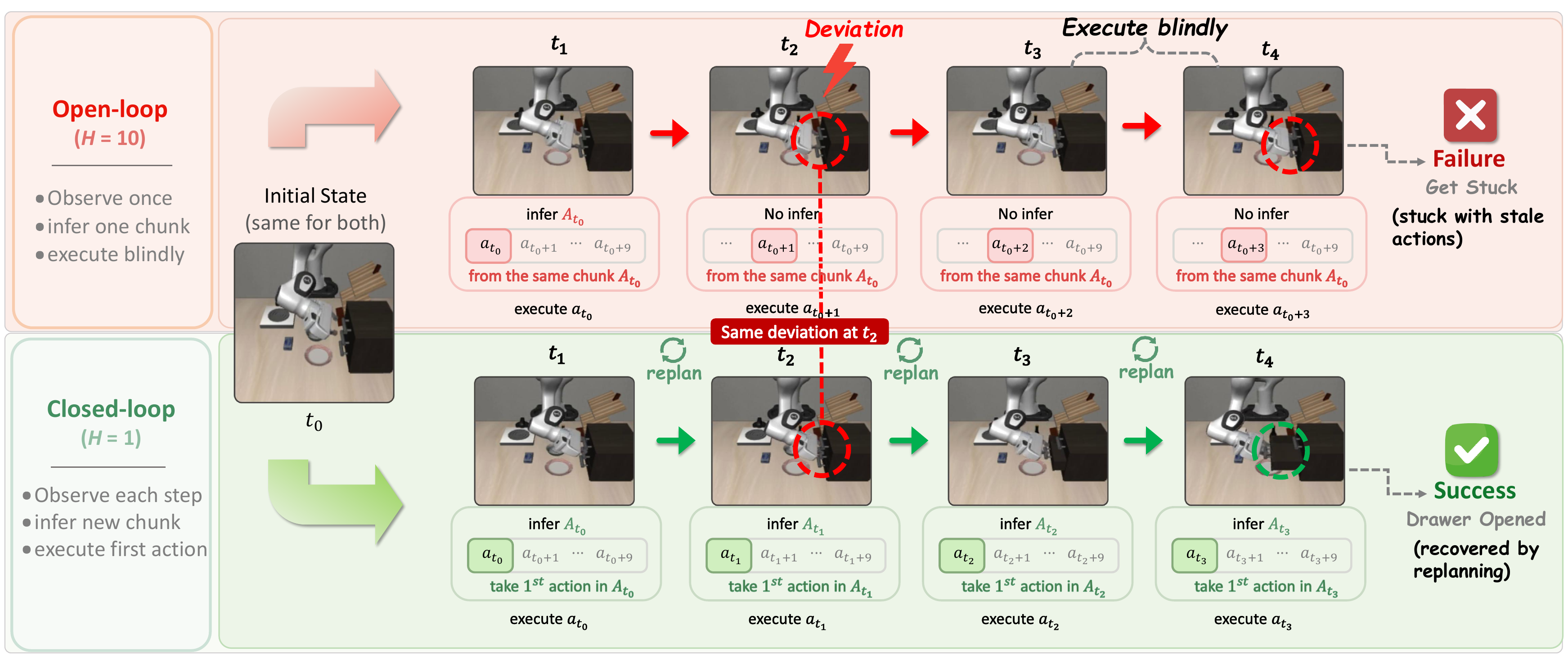}
    \caption{
         \textbf{Open-loop vs. Closed-loop execution.}
         The top row ($H=10$) illustrates how blindly executing long action chunks leads to compounding errors, causing the robot to get stuck during a drawer-opening task. In contrast, the bottom row ($H=1$) demonstrates strict closed-loop execution, which maintains environmental reactivity and yields a smooth, successful manipulation.
    }
    \label{fig:open_loop_vs_closed_loop}
\end{figure}

A common engineering compromise is to use an \textbf{action chunk}~\cite{zhao2023learning}: in a single forward pass, the policy predicts a sequence of future actions, while the controller only executes the first $H$ steps as the \textbf{action horizon}~\cite{jing2025mixture,liu2024bidirectional,sendai2025leave,chi2025diffusion,khan2025test,so2025improving}. This design amortizes policy-call frequency and improves temporal smoothness, but it also creates an \textbf{open-loop blind spot}~\cite{wang2026open,xu2026vgas} that introduces two compounding risks. \textbf{First, the policy lacks real-time reactivity}. Fresh observations arrive at every control step, yet the system ignores them until the horizon ends, leaving it unable to respond to unexpected slippage, collision, or pose drift as they occur. \textbf{Second, errors can accumulate beyond the point that replanning can recover from during the blind spot}. If deviations go uncorrected long enough, the robot may drift into an out-of-distribution state that the policy has rarely encountered during training, at which point even the next replanning call cannot steer execution back to the intended trajectory~\cite{wang2026open,liu2024bidirectional,sendai2025leave,wu2026closed,xuvla}, and the task ultimately fails. Crucially, both risks worsen as the horizon grows longer. As illustrated in Fig.~\ref{fig:open_loop_vs_closed_loop}, under a long horizon ($H=10$), the robot keeps executing stale actions after a deviation and eventually gets stuck; in contrast, with $H=1$, the policy replans at every step and avoids the same error accumulation.

\begin{wrapfigure}{r}{0.45\textwidth}
    \vspace{0pt}
    \centering
    \includegraphics[width=\linewidth]{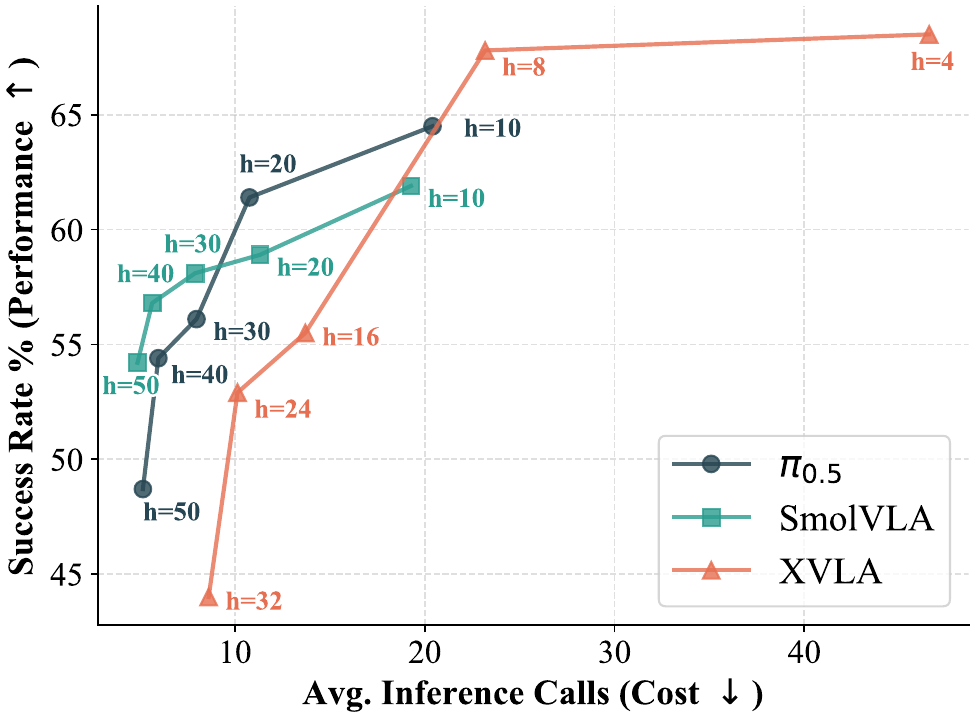}
    \caption{
    \textbf{Performance--efficiency trade-off across fixed action horizons.}
    Smaller horizon achieves higher success rates, while larger horizon preserves the chunking efficiency.
    }
    \label{fig:teaser_pareto}
    \vspace{0pt}
\end{wrapfigure}

Yet $H=1$ is not a practical solution: it requires a full VLA inference at every control step, negating the efficiency gains that action chunking was designed to provide. The contrast between these two extremes thus exposes a fundamental trade-off: smaller horizons improve closed-loop responsiveness but multiply policy-call frequency, while larger horizons amortize computation but widen the blind spot where errors silently accumulate.

To quantify this trade-off, we conduct systematic experiments comparing task success rate and policy frequency across a range of action horizons on three VLA backbones. As shown in Fig.~\ref{fig:teaser_pareto}, larger horizons substantially reduce policy calls but also lead to lower success rates, and this trend is consistently observed across all three backbones. For $\pi_{0.5}$~\cite{intelligence2025pi_}, for example, increasing the horizon reduces policy calls by about $4\times$, but lowers success from about 64\% to below 49\%; similar trends hold for SmolVLA~\cite{shukor2025smolvla} and X-VLA~\cite{zheng2025x}. Furthermore, since the best horizon depends on task difficulty, environmental dynamics, and sim-to-real mismatch, it is difficult to predefine a single static horizon that remains optimal across scenarios. This suggests that the key is not to choose a better fixed horizon, but to decide when the current chunk should stop being trusted. 

The above analysis raises two key questions: (1) How to \textbf{detect execution deviations in time} and \textbf{terminate stale actions} before errors compound beyond recovery;  (2) How to \textbf{correct the trajectory after truncation}---since naive replanning alone is often insufficient: the VLA may re-generate actions that still fail to escape the deviated state, leaving the robot trapped again.

Motivated by these insights, we propose \textbf{VLA-Corrector}, a lightweight corrective inference framework that addresses both questions. For the first, VLA-Corrector introduces a \textbf{Latent-space Vision Monitor} (LVM) that continuously compares predicted and actual visual evolution during open-loop execution; once persistent deviation is detected, it triggers an \textbf{interrupt event} that immediately truncates the remaining stale actions and invokes a new action generation. For the second, rather than relying on naive replanning, VLA-Corrector activates \textbf{Online Gradient Guidance} (OGG), which exploits the discrepancy between predicted and observed visual dynamics to inject a corrective gradient during new action generation---actively steering the robot back toward the intended trajectory, instead of simply hoping that re-inference will naturally recover. Together, LVM-triggered truncation and OGG-guided re-inference transform a fixed action horizon into an \textbf{adaptive action horizon} with \textbf{self-correcting re-inference}, preserving long-horizon efficiency and short-horizon responsiveness.

These two mechanisms jointly improve success-per-call efficiency: the robot avoids wasting long chunks on stale actions, so each policy query contributes more effectively to task completion. For $\pi_{0.5}$ at horizon 50, success rises from 48.7\% to 58.7\% while average policy calls drop from 5.15 to 4.98---a $+24.6\%$ success-per-call efficiency gain. Gains are consistent across all three backbones and grow larger at longer horizons, where the open-loop blind spot is widest. Beyond robustness, VLA-Corrector also improves sample efficiency: on LIBERO, a few-shot fine-tuned model augmented with VLA-Corrector reaches 97.8\% average success, \emph{surpassing} the fully fine-tuned baseline (96.9\%).

Our contributions are summarized as follows:

\textbullet~We systematically quantify the performance--efficiency trade-off induced by fixed action horizon in action-chunked VLA policies, showing that the open-loop blind spot consistently affects robustness across different VLA backbones.

\textbullet~We propose VLA-Corrector, a lightweight inference-time framework that augments a frozen VLA backbone with latent-space monitoring, event-triggered truncation, and guidance toward recovery-oriented inference.

\textbullet~We validate VLA-Corrector across simulation and real-world tasks, showing that it achieves stronger robustness, higher success-per-call efficiency, and adaptive correction behavior.

\section{Preliminaries}

\textbf{VLA Policies with Action Chunks.}
In embodied continuous control, the current visual observation $o_t$ is first encoded into a latent representation $Z_t^{\mathrm{real}} = \mathcal{E}(o_t)$ by a visual encoder $\mathcal{E}$. To balance policy-call frequency and action smoothness, modern generative VLA policies parameterized by $\theta$ commonly predict an \textbf{action chunk} in a single inference call,
\begin{equation}
A_t = [a_t, a_{t+1}, \dots, a_{t+C-1}] \sim \pi_\theta(\cdot \mid Z_t^{\mathrm{real}}, l),
\label{eq:action_chunk}
\end{equation}
where $a_t$ denotes a single control action and $l$ is the language instruction. During deployment, only the first $H$ actions, with $H \leq C$, are executed sequentially, and we refer to this execution window as the \emph{action horizon}. The corresponding execution queue is $Q_t = [a_t, a_{t+1}, \dots, a_{t+H-1}]$, during which the controller does not query the VLA policy again.


\section{VLA-Corrector: A Lightweight Corrective Inference Framework}
\begin{figure}[t]
    \centering
    \includegraphics[width=\textwidth]{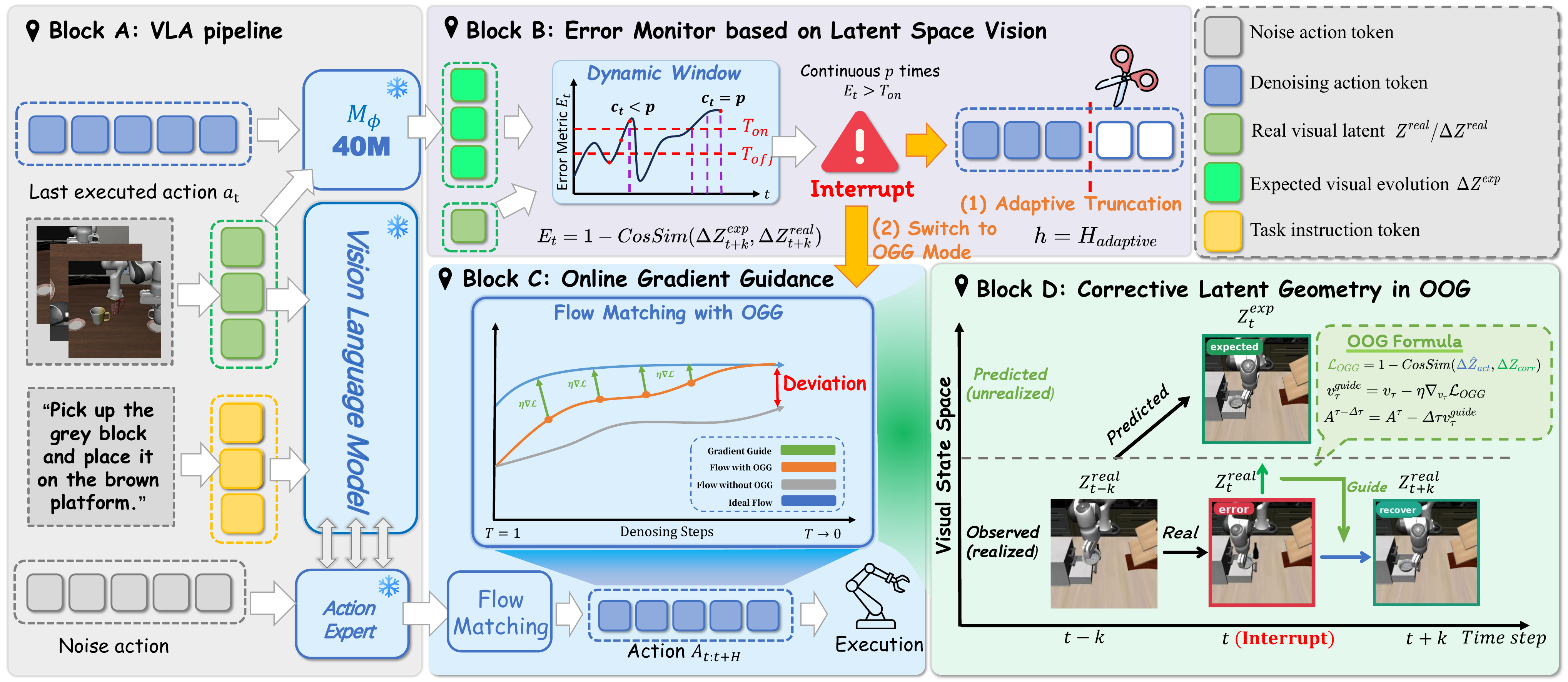}
    \caption{
        \textbf{Overview of VLA-Corrector.}
        Starting from a standard chunked VLA pipeline (\textbf{A}), we add a \textbf{Latent-space Vision Monitor} (LVM) that detects persistent execution drift and triggers an interrupt event (\textbf{B}). The event truncates stale actions and switches the next replan from normal flow matching to OGG-guided flow matching (\textbf{C}). OGG uses the expected and observed latent evolution to guide the replan back toward a recoverable trajectory (\textbf{D}).
    }
    \label{fig:method_overview}
\end{figure}
VLA-Corrector addresses the open-loop blind spot of action-chunked VLA policies. During the action horizon, fresh observations are available, but the policy does not use them until the horizon ends. We therefore decouple \emph{action generation} from \emph{execution monitoring}: the VLA backbone generates action chunks, while VLA-Corrector monitors whether execution remains on track and intervenes only when drift emerges. We organize the method as follows: \textbf{(1) Corrector Training} describes the external latent dynamics corrector; \textbf{(2) LVM Detection} introduces online visual deviation detection; \textbf{(3) Event-Triggered Truncation} explains how persistent deviations induce adaptive horizons; and \textbf{(4) OGG Correction} presents corrective replanning after truncation.

\subsection{Training the External Latent Dynamics Corrector} VLA-Corrector is trained after a VLA policy has already been obtained. We first fine-tune the VLA backbone on the benchmark training set, then freeze the VLA and use its visual encoder $\mathcal{E}$ to extract visual latents from demonstration trajectories. Given a transition $(o_t,a_t,o_{t+k})$, we compute \begin{equation} Z_t^{\mathrm{real}}=\mathcal{E}(o_t), \qquad Z_{t+k}^{\mathrm{real}}=\mathcal{E}(o_{t+k}), \qquad \Delta Z_{t+k}^{*}=Z_{t+k}^{\mathrm{real}}-Z_t^{\mathrm{real}}, \label{eq:corrector_target} \end{equation} where $\Delta Z_{t+k}^{*}$ is the target short-horizon visual latent evolution induced by the executed action. We then train a lightweight external dynamics corrector $M_\phi$ to predict this residual evolution from the current latent state and action: \begin{equation} \Delta \hat{Z}_{t+k}=M_\phi(Z_t^{\mathrm{real}},a_t). \label{eq:corrector_prediction} \end{equation} Rather than predicting the absolute future latent state, $M_{\phi}$ instead predicts a residual visual latent evolution, which actively suppresses static scene content and encourages the model to focus on task-relevant dynamics. The training objective combines magnitude matching and directional consistency: \begin{equation} \mathcal{L}_{\mathrm{corr}} = \left\| \Delta \hat{Z}_{t+k}-\Delta Z_{t+k}^{*} \right\|_2^2 + \beta \left[ 1- \mathrm{CosSim} \left( \Delta \hat{Z}_{t+k}, \Delta Z_{t+k}^{*} \right) \right], \label{eq:corrector_loss} \end{equation} where $\beta$ balances residual accuracy and directional alignment. The trainable component is therefore not the VLA policy itself, but a lightweight latent dynamics module trained separately on top of frozen VLA features. This modular design allows the corrector to be retrained or replaced for each benchmark without re-optimizing the expensive VLA backbone. The corrector is trained on benchmark demonstration trajectories because they are easy to obtain and broadly reflect successful task progress. Its goal is not to model all possible futures as a full world model would do, but to learn whether local latent dynamics remain consistent with on-track execution. Demonstrations are suitable for this purpose: although they may contain small teleoperation jitter or local imperfections, they still capture behavior that keeps the task progressing correctly. Since this is a local and low-dimensional prediction task, a lightweight MLP is sufficient in practice; in our experiments, a 40M corrector already provides an effective monitoring and corrective signal, which can be trained at a very low cost.

\subsection{Latent Visual Dynamics for Online Anomaly Detection}

After training $M_\phi$, we use it online to monitor whether the current action chunk remains reliable. During execution, \textbf{Latent-space Vision Monitor} (LVM) compares the \emph{expected} latent visual evolution predicted by $M_\phi$ with the \emph{actual} evolution observed from fresh camera input, detecting deviations before they accumulate into failure.

\textbf{Expected residual dynamics.}
At control step $t$, LVM uses the executed action $a_t$ and current latent state $Z_t^{\mathrm{real}}$ to predict the expected short-horizon latent residual, $\Delta Z_{t+k}^{\mathrm{exp}}=M_\phi(Z_t^{\mathrm{real}},a_t)$, where $k$ is the prediction interval.

\textbf{Online inconsistency score.}
During rollout, the latest observation remains available even though the policy is not re-queried. We encode the future observation and compute the actual residual $\Delta Z_{t+k}^{\mathrm{real}}=Z_{t+k}^{\mathrm{real}}-Z_t^{\mathrm{real}}$. LVM then measures the mismatch between expected and actual latent evolution:
\begin{equation}
E_t = 1 - \mathrm{CosSim}\!\left(\Delta Z_{t+k}^{\mathrm{exp}}, \Delta Z_{t+k}^{\mathrm{real}}\right),
\qquad
\mathrm{CosSim}(\mathbf{u},\mathbf{v}) =
\mathbf{u}^{\top}\mathbf{v}/(\|\mathbf{u}\|\,\|\mathbf{v}\|).
\label{eq:inconsistency_score}
\end{equation}
A larger $E_t$ indicates stronger visual dynamics mismatch, providing a continuous signal for later event-triggered truncation.

\subsection{Event-Triggered Truncation under Robust Online Monitoring}

Given the continuous inconsistency score $E_t$, we need to decide when a deviation is persistent enough to justify intervention. Directly thresholding $E_t$ can be unstable, since transient visual outliers may cause false triggers. We therefore use a robust event-triggered rule based on dynamic thresholds and persistence checking. Specifically, we maintain a sliding window of recent scores, $\mathbf{E}_W=\{E_{t-w+1},\dots,E_t\}$, and compute its median $M_e$ and median absolute deviation,
\begin{equation}
\mathrm{MAD}=\mathrm{median}\!\left(|E_i-M_e|\right), \qquad E_i\in\mathbf{E}_W .
\label{eq:mad_main}
\end{equation}
We then define two adaptive thresholds,
\begin{equation}
T_{\mathrm{on}} = M_e+\lambda_{\mathrm{on}}\mathrm{MAD}, 
\qquad
T_{\mathrm{off}} = M_e+\lambda_{\mathrm{off}}\mathrm{MAD},
\qquad
\lambda_{\mathrm{on}}>\lambda_{\mathrm{off}},
\label{eq:dual_threshold_main}
\end{equation}
where $T_{\mathrm{on}}$ confirms persistent deviation and $T_{\mathrm{off}}$ provides hysteresis for recovery. An interrupt event is triggered only when $E_t>T_{\mathrm{on}}$ holds for $p$ consecutive steps; isolated spikes are ignored.

Once an interrupt is triggered, VLA-Corrector discards the remaining actions in the current queue and queries the policy again under corrective mode. If the controller has executed $h$ actions from the original queue, the realized horizon becomes $H_{\mathrm{adaptive}}=h<H$. Thus, the fixed action horizon is shortened only when sustained visual drift indicates that the current chunk has become stale. The full thresholding state machine is provided in Appendix~\ref{app:event_trigger_details}.

\subsection{Online Gradient Guidance for Corrective Inference}

Truncation stops stale actions, but recovery still depends on the next replan. After an interrupt event, VLA-Corrector applies \textbf{Online Gradient Guidance} (OGG)~\cite{park2025acg} only to the single policy call immediately following the interrupt event, guiding this recovery replan toward a corrective latent direction.

\textbf{Predicted action effect.}
At denoising step $\tau$ of flow matching, let $A^\tau$ be the noisy action chunk. The VLA predicts a velocity field $v_\tau=\pi_\theta(A^\tau,Z_t^{\mathrm{real}},\tau)$, from which we estimate the clean chunk $\hat{A}_0=A^\tau-\tau v_\tau$ and take its first action $\hat{a}_t=\hat{A}_0[0]$. The corrector then predicts the latent effect of this candidate action:
\begin{equation}
\Delta\hat{Z}_{\mathrm{act}}=M_\phi(Z_t^{\mathrm{real}},\hat{a}_t).
\label{eq:ogg_predicted_effect}
\end{equation}

\textbf{Corrective target.}
Let $t-k$ be the last stable step before an interrupt event. We define the expected residual $\Delta Z_{\mathrm{exp}} = \Delta Z_t^{\mathrm{exp}} = M_\phi(Z_{t-k}^{\mathrm{real}}, a_{t-k})$, which is predicted by $M_\phi$ from step $t-k$. The accumulated deviation is $\Delta Z_{\mathrm{dev}} = \Delta Z_t^{*} = Z_t^{\mathrm{real}} - Z_{t-k}^{\mathrm{real}}$. The corrective latent direction is
\begin{equation}
\Delta Z_{\mathrm{corr}}=\Delta Z_{\mathrm{exp}}-\Delta Z_{\mathrm{dev}}.
\label{eq:corrective_direction_compact}
\end{equation}
This direction preserves the intended local dynamics while compensating for the drift accumulated during open-loop execution.

\textbf{Guided velocity update.}
OGG aligns the predicted action effect with $\Delta Z_{\mathrm{corr}}$ through
\begin{equation}
\mathcal{L}_{\mathrm{OGG}}=1-\mathrm{CosSim}(\Delta\hat{Z}_{\mathrm{act}},\Delta Z_{\mathrm{corr}}),
\label{eq:ogg_loss_compact}
\end{equation}
and injects the resulting gradient into the flow-matching velocity:
\begin{equation}
v_\tau^{\mathrm{guide}}=v_\tau-\eta\nabla_{v_\tau}\mathcal{L}_{\mathrm{OGG}},
\qquad
A^{\tau-\Delta\tau}=A^\tau-\Delta\tau v_\tau^{\mathrm{guide}} .
\label{eq:ogg_guidance_compact}
\end{equation}
Here, $\eta$ controls the guidance strength. Since OGG modifies the velocity field rather than directly perturbing action coordinates, it remains compatible with the original flow-matching process and yields smoother corrective replanning.

\section{Experiments}
We evaluate VLA-Corrector on MetaWorld~\cite{yu2020meta}, LIBERO~\cite{liu2023libero}, RoboTwin v2.0~\cite{chen2025robotwin}, and a real AgileX PiPER robot, using $\pi_{0.5}$~\cite{intelligence2025pi_} as the main backbone and SmolVLA~\cite{shukor2025smolvla}/X-VLA~\cite{zheng2025x} for cross-architecture evaluation. The experiments cover four aspects: \textbf{(1) Performance and policy-call efficiency}, evaluating success rate and success-per-call across benchmarks and horizons; \textbf{(2) Mechanism analysis}, examining LVM detection, truncation timing, and OGG recovery; \textbf{(3) Real-world transfer}, testing the method on physical robot manipulation; and \textbf{(4) Ablations}, analyzing component necessity and sensitivity. Detailed protocols, metrics, and compute resources are provided in Appendix~\ref{app:benchmark_protocol}.

\subsection{Main Results}

\begin{table*}[t]
  \centering
  \small
  \setlength{\tabcolsep}{4.5pt}
  \renewcommand{\arraystretch}{1.12}
  \caption{\textbf{Cross-architecture generalization on MetaWorld.}
  Success rate (\%, $\uparrow$) across difficulty splits. Green arrows show absolute improvements over the corresponding baseline.}
  \label{tab:metaworld_generalization}
  \begin{tabular}{llccccc}
    \toprule
    \rowcolor{headerbg}
    \textbf{Backbone} & \textbf{Method} & \textbf{Easy}$\uparrow$ & \textbf{Medium}$\uparrow$ & \textbf{Hard}$\uparrow$ & \textbf{Very Hard}$\uparrow$ & \textbf{Avg.}$\uparrow$ \\
    \midrule

    \multirow{2}{*}{$\pi_{0.5}$}
    & Baseline & 70.5 & 45.0 & 38.3 & 41.0 & 48.70 \\
    & \cellcolor{oursbg}\textbf{+ VLA-Corrector}
      & \cellcolor{oursbg}\textbf{83.2} \gain{12.7}
      & \cellcolor{oursbg}\textbf{61.7} \gain{16.7}
      & \cellcolor{oursbg}\textbf{47.5} \gain{9.2}
      & \cellcolor{oursbg}\textbf{65.0} \gain{24.0}
      & \cellcolor{oursbg}\textbf{64.35} \gain{15.65} \\
    \midrule

    \multirow{2}{*}{SmolVLA}
    & Baseline & 81.3 & 53.6 & 51.7 & 61.0 & 61.90 \\
    & \cellcolor{oursbg}\textbf{+ VLA-Corrector}
      & \cellcolor{oursbg}\textbf{83.4} \gain{2.1}
      & \cellcolor{oursbg}\textbf{56.0} \gain{2.4}
      & \cellcolor{oursbg}\textbf{64.2} \gain{12.5}
      & \cellcolor{oursbg}\textbf{63.0} \gain{2.0}
      & \cellcolor{oursbg}\textbf{66.65} \gain{4.75} \\
    \midrule

    \multirow{2}{*}{X-VLA}
    & Baseline & 72.5 & 46.4 & 48.3 & 55.0 & 55.55 \\
    & \cellcolor{oursbg}\textbf{+ VLA-Corrector}
      & \cellcolor{oursbg}\textbf{74.4} \gain{1.9}
      & \cellcolor{oursbg}\textbf{50.0} \gain{3.6}
      & \cellcolor{oursbg}\textbf{50.0} \gain{1.7}
      & \cellcolor{oursbg}\textbf{64.0} \gain{9.0}
      & \cellcolor{oursbg}\textbf{59.60} \gain{4.05} \\
    \bottomrule
  \end{tabular}
\end{table*}

\textbf{Cross-Architecture Evaluation on MetaWorld.}
VLA-Corrector consistently improves all three VLA backbones on MetaWorld, with larger gains on harder tasks. As shown in Table~\ref{tab:metaworld_generalization}, the average success rate improves by 15.65 points for $\pi_{0.5}$, 4.75 points for SmolVLA, and 4.05 points for X-VLA. The strongest gain appears on the Very Hard split of $\pi_{0.5}$, increasing success from 41.0\% to 65.0\%.

\begin{table*}[t]
  \centering
  \small
  \setlength{\tabcolsep}{4.7pt}
  \renewcommand{\arraystretch}{1.12}
  \caption{\textbf{RoboTwin v2.0 evaluation.}
  Success rate (\%, $\uparrow$), average policy calls per episode ($\downarrow$), and per-step wall-clock latency ($\downarrow$). The benchmark covers all 50 tasks in clean and randomized configurations, with 50 episodes per task/configuration (5,000 episodes per method); the same frozen $\pi_{0.5}$ backbone is used throughout.}
  \label{tab:robotwin_main}
  \begin{tabular}{lccccc}
    \toprule
    \rowcolor{headerbg}
    \textbf{Method} & \textbf{Clean}$\uparrow$ & \textbf{Random}$\uparrow$ & \textbf{Overall (95\% CI)}$\uparrow$ & \textbf{Calls/ep.}$\downarrow$ & \textbf{ms/step}$\downarrow$ \\
    \midrule
    Baseline $H=50$ & 68.2 & 62.6 & 65.4 [64.1,66.7] & 5.74 & 8.10 \\
    Baseline $H=30$ (latency-matched) & 72.0 & 66.8 & 69.4 [68.1,70.7] & 9.41 & 9.98 \\
    \rowcolor{oursbg}
    \textbf{VLA-Corrector $H=50$} & \textbf{76.0} & \textbf{70.2} & \textbf{73.1 [71.9,74.3]} & 6.88 & 10.02 \\
    \bottomrule
  \end{tabular}
\end{table*}

\textbf{RoboTwin Evaluation.}
On the broader RoboTwin v2.0 benchmark, VLA-Corrector improves overall success from 65.4\% to 73.1\% over the original $H=50$ baseline and from 69.4\% to 73.1\% over the latency-matched $H=30$ baseline. The gain over the matched baseline is 3.7 points while per-step latency remains nearly identical (10.02 vs. 9.98~ms); VLA-Corrector also uses 27\% fewer policy calls than the matched baseline (6.88 vs. 9.41 calls/episode). The improvement persists in both clean and randomized configurations. A more comprehensive and detailed analysis can be found in Appendix~\ref{app:async_rtc_matched}.

\begin{table*}[t]
  \centering
  \small
  \setlength{\tabcolsep}{5.2pt}
  \renewcommand{\arraystretch}{1.12}
  \caption{\textbf{Sample efficiency on LIBERO.}
  Success rate (\%, $\uparrow$). Green arrows compare VLA-Corrector against the few-shot fine-tuned baseline.}
  \label{tab:libero_efficiency}
  \begin{tabular}{lccccc}
    \toprule
    \rowcolor{headerbg}
    \textbf{Model} & \textbf{Object}$\uparrow$ & \textbf{Spatial}$\uparrow$ & \textbf{Goal}$\uparrow$ & \textbf{Long}$\uparrow$ & \textbf{Avg.}$\uparrow$ \\
    \midrule
    $\pi_{0.5}$ (Full Fine-tuned) & 99.4 & 98.2 & 97.8 & 92.4 & 96.95 \\
    \midrule
    $\pi_{0.5}$ (Few-shot Fine-tuned) & 97.8 & 95.4 & 96.2 & 86.6 & 94.00 \\
    \rowcolor{oursbg}
    $\pi_{0.5}$ (Few-shot) \textbf{+ VLA-Corrector}
      & \textbf{99.8} \gain{2.0}
      & \textbf{100.0} \gain{4.6}
      & \textbf{98.0} \gain{1.8}
      & \textbf{93.4} \gain{6.8}
      & \textbf{97.80} \gain{3.80} \\
    \bottomrule
  \end{tabular}
\end{table*}

\textbf{Sample Efficiency on LIBERO.}
We further evaluate whether VLA-Corrector can compensate for limited task-specific data.
Table~\ref{tab:libero_efficiency} compares a fully fine-tuned $\pi_{0.5}$ model with a few-shot fine-tuned counterpart on LIBERO.
Starting from the publicly released LeRobot~\cite{cadene2026lerobot} few-shot checkpoint 
\texttt{pi05\_libero\_base}, VLA-Corrector improves the average success rate from 94.00\% to 97.80\%, even surpassing the fully fine-tuned baseline (96.95\%).

These results suggest that few-shot fine-tuning can already learn much of the normal task trajectories, but may lack coverage of drifted states and their recovery behaviors.
Instead of improving recovery by exposing the backbone to more rare failure cases during training, VLA-Corrector makes recovery easier at inference time by interrupting error accumulation early and guiding corrective replanning, thereby alleviating the dependence on additional post-training data for robust recovery.

\begin{figure*}[t]
    \centering
    \begin{subfigure}[b]{0.48\textwidth}
        \centering
        \includegraphics[height=0.16\textheight,keepaspectratio]{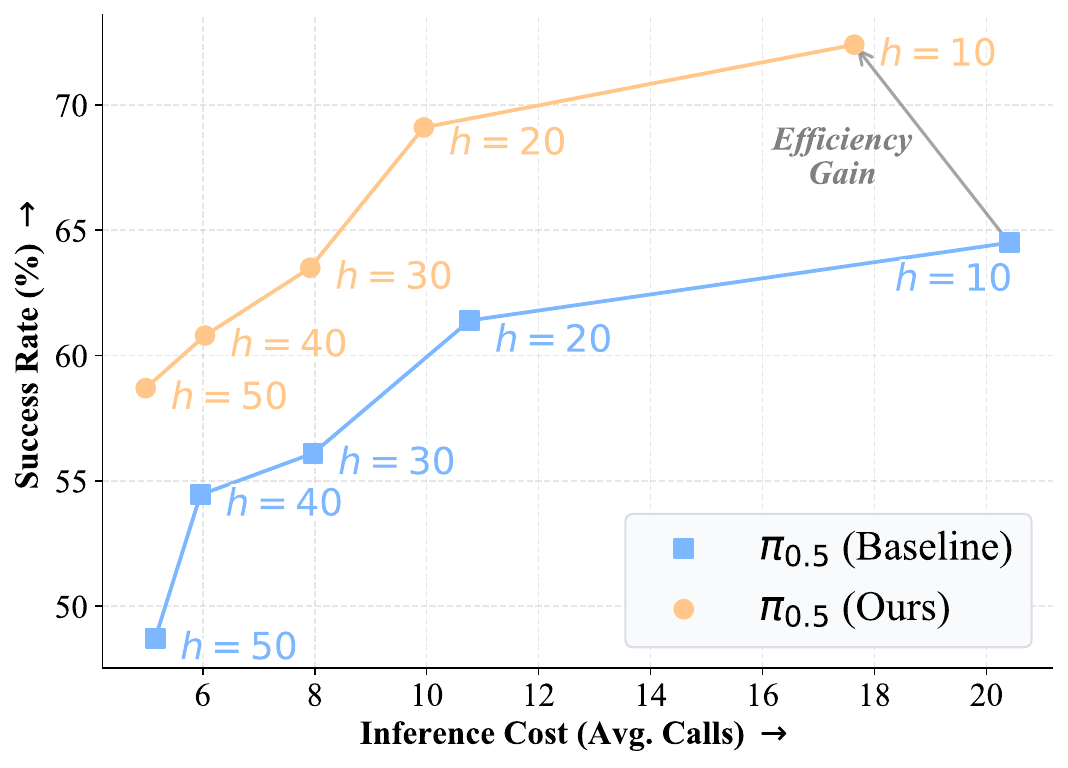}
    \end{subfigure}
    \hfill
    \begin{subfigure}[b]{0.48\textwidth}
        \centering
        \includegraphics[height=0.16\textheight,keepaspectratio]{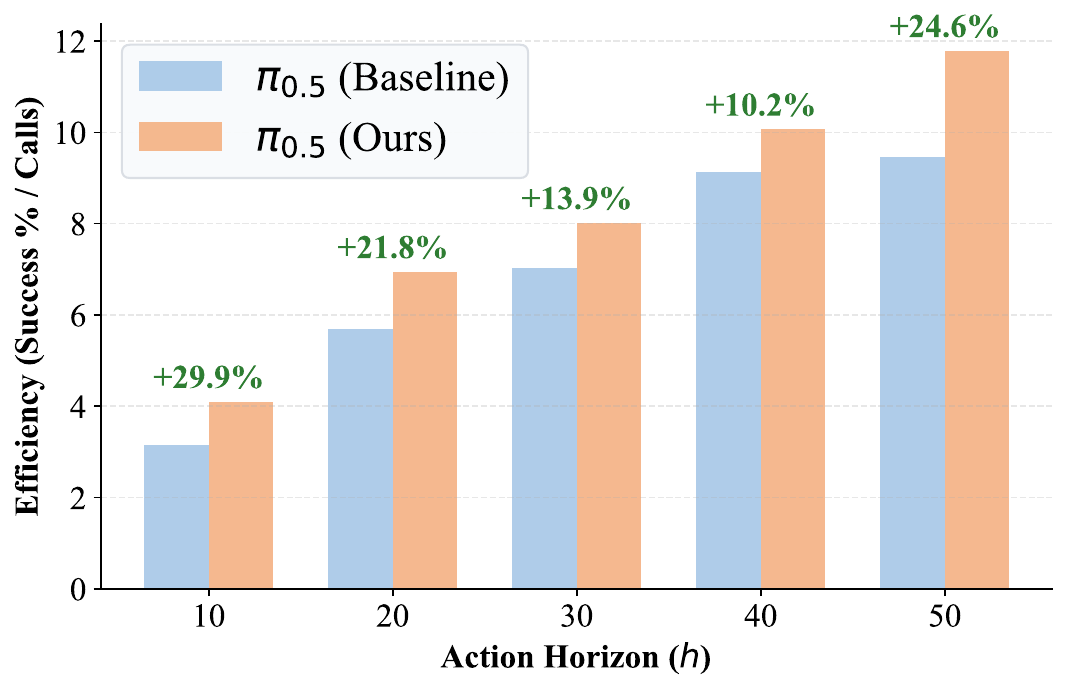}
    \end{subfigure}
    \caption{
    \textbf{Performance--efficiency analysis on $\pi_{0.5}$.}
    \textbf{Left}: performance--efficiency trade-off across action horizons.
    \textbf{Right}: success-per-call efficiency.
    VLA-Corrector improves success rate across action horizons and yields consistent efficiency gains.
    }
    \label{fig:pi05_efficiency}
\end{figure*}

\textbf{Performance--Efficiency Trade-off.}
VLA-Corrector improves success-per-call efficiency across action horizons rather than simply querying the policy more often. As shown in Figure~\ref{fig:pi05_efficiency}, the largest success-per-call gains reach 29.9\% for $\pi_{0.5}$, 45.3\% for SmolVLA, and 39.1\% for X-VLA. For example, on SmolVLA with horizon 10, success improves from 61.90\% to 73.00\% while policy calls decrease from 19.27 to 15.64. This suggests that VLA-Corrector makes each policy query more useful by interrupting stale chunks and guiding recovery. Full results are provided in Appendix~\ref{app:full_efficiency}.

\subsection{Mechanism Analysis}
In this section, we will analyze the mechanisms behind VLA-Corrector from three perspectives: 
\textbf{1) LVM detection}, whether LVM provides a useful and well-timed drift signal; and 
\textbf{2) OGG correction}, whether OGG improves recovery after an interrupt event;
\textbf{3) controlled recovery}, whether the full framework improves recovery in a controlled failure case.

\begin{figure*}[htbp]
    \centering
    \setlength{\abovecaptionskip}{2pt}

    \begin{subfigure}[t]{0.66\textwidth}
        \centering
        \includegraphics[width=\linewidth]{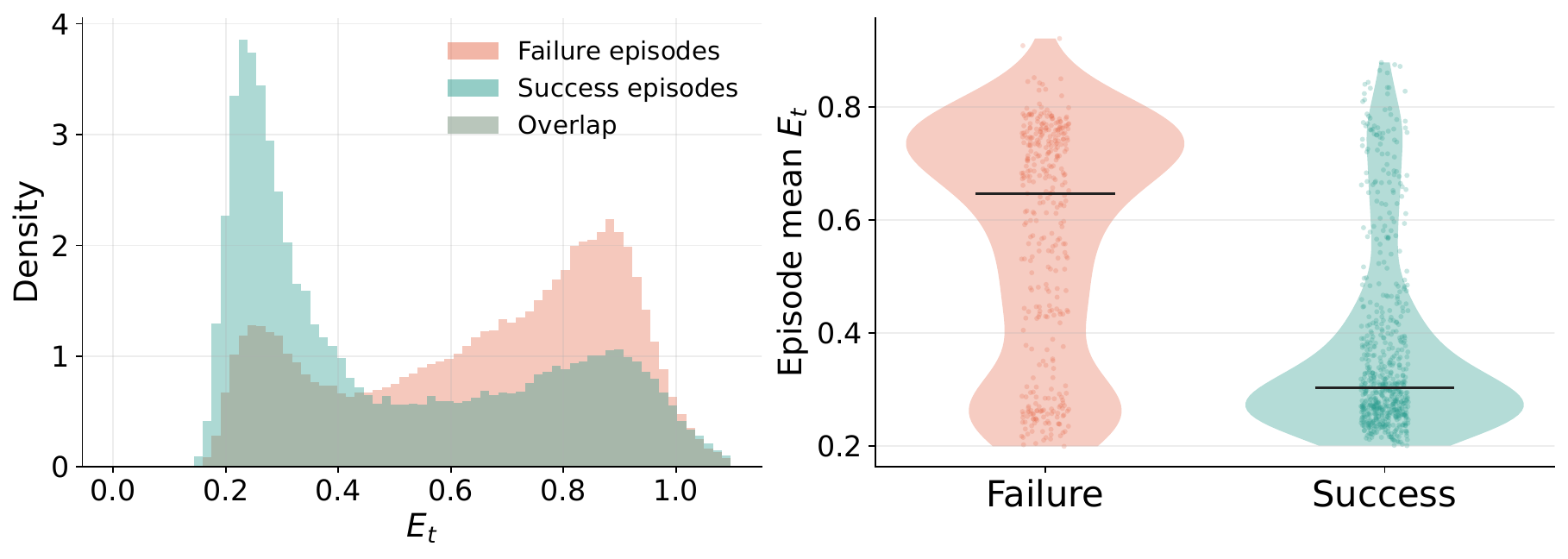}
        \label{fig:et_density}
    \end{subfigure}
    \hfill
    \begin{subfigure}[t]{0.32\textwidth}
        \centering
        \includegraphics[width=\linewidth]{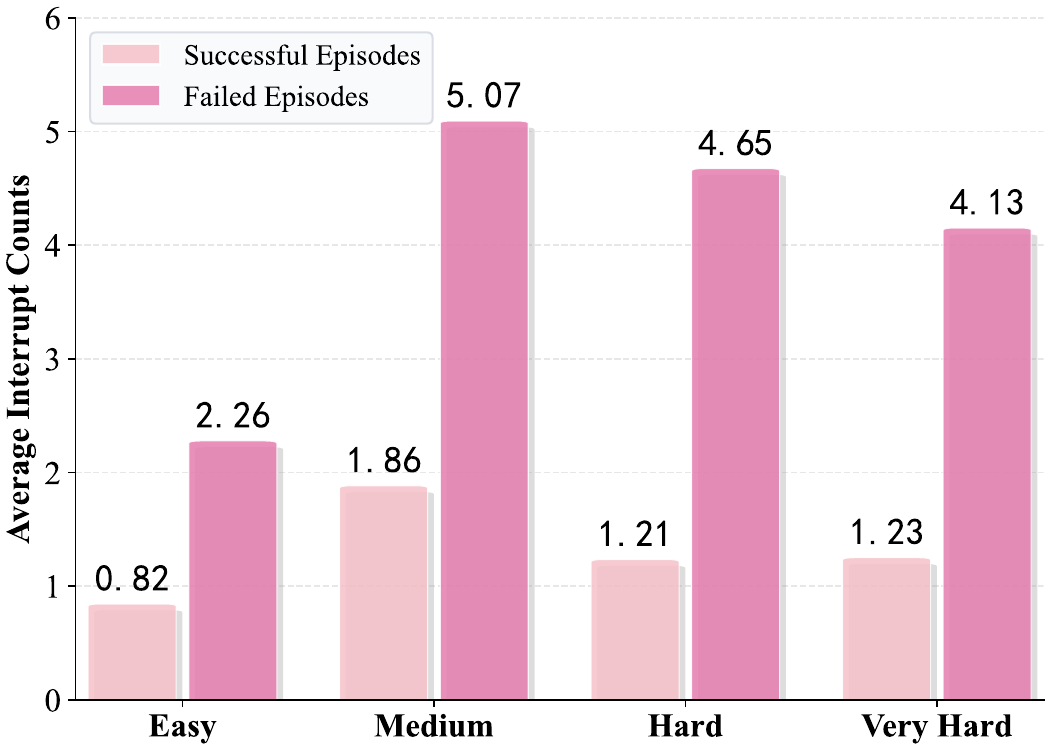}
        \label{fig:detection_analysis}
    \end{subfigure}

    \caption{
    \textbf{LVM detection analysis.}
    \textbf{Left}: distribution of the inconsistency score $E_t$, where successful episodes concentrate at low values and failed episodes show a heavier high-score tail.
    \textbf{Right}: interrupt frequency, where failed episodes trigger more interrupt events than successful ones.
    }
    \label{fig:lvm_detection_analysis}
\end{figure*}

\textbf{LVM Detection.}
LVM provides an informative drift signal. As shown in Figure~\ref{fig:lvm_detection_analysis}, successful episodes concentrate at low $E_t$, while failed episodes show a heavier high-score tail and trigger more interrupt events. This indicates that $E_t$ captures failure-prone visual dynamics mismatch.

\begin{figure*}[hbtp]
    \centering

    \begin{minipage}[t]{0.66\textwidth}
        \vspace{0pt}
        \centering
        \includegraphics[width=\linewidth]{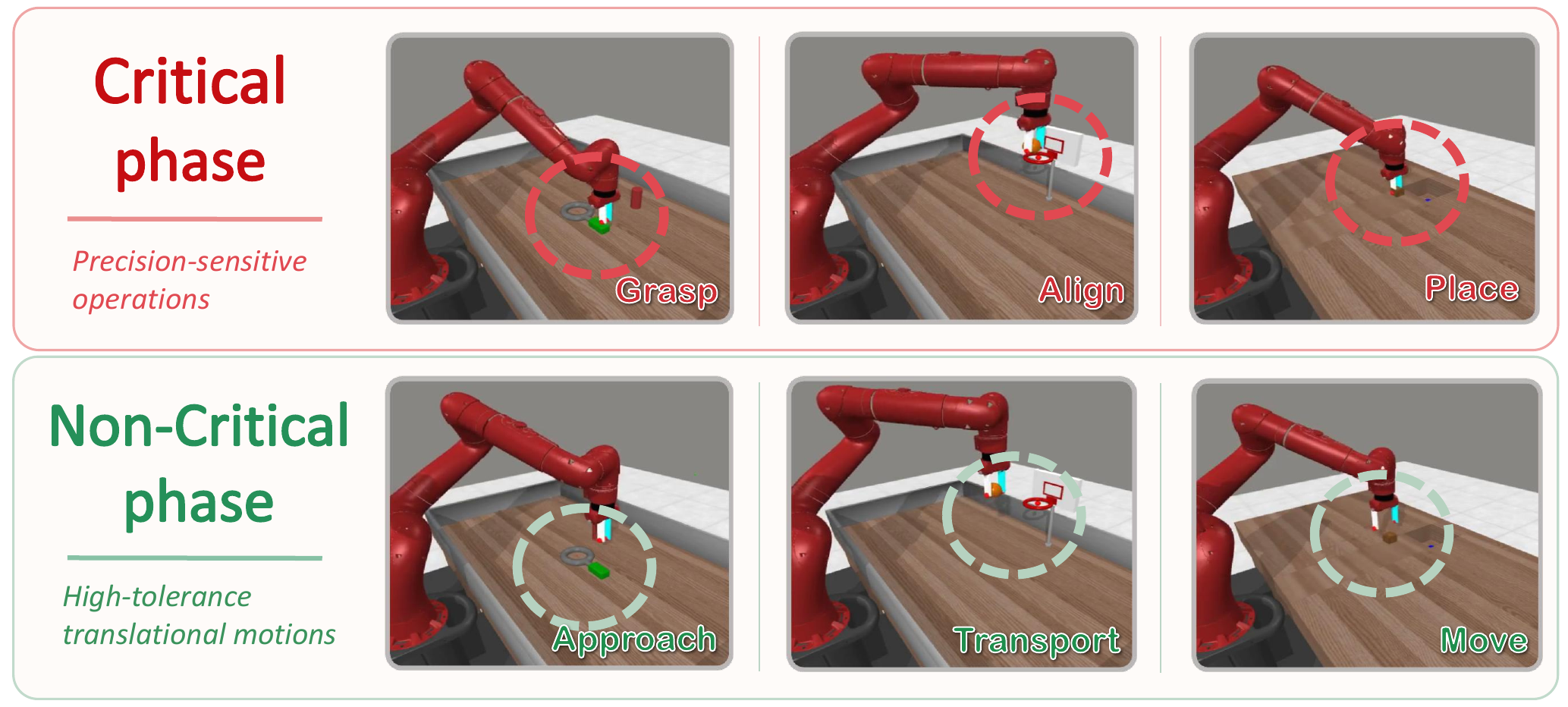}
    \end{minipage}
    \hfill
    \begin{minipage}[t]{0.28\textwidth}
        \vspace{0pt}
        \centering
        \small
        \renewcommand{\arraystretch}{1.25}
        \setlength{\tabcolsep}{3pt}

        \fboxsep=6pt
        \colorbox{white}{%
        \begin{minipage}{\dimexpr\linewidth-2\fboxsep\relax}
            \centering
            {\bfseries Truncation Distribution}\\[-0.1em]
            {\bfseries by Task Phase}

            \vspace{0pt}

            \begin{tabular}{lc}
                \toprule
                \rowcolor{groupbg}
                \textbf{Phase} & \textbf{Rate} \\
                \midrule
                \rowcolor{softgreen}
                Critical & \textbf{83.7\%} \\
                Non-critical & 16.3\% \\
                \bottomrule
            \end{tabular}

            \vspace{0pt}

            \begin{tabular}{p{\dimexpr\linewidth-2\tabcolsep\relax}}
                \rowcolor{softgreen}
                \centering
                \textbf{5.1$\times$ more truncations occur in critical phases than in non-critical phases.}
            \end{tabular}
        \end{minipage}%
        }
    \end{minipage}

    \caption{
    \textbf{Task-phase analysis of LVM-triggered truncation.}
    We manually divide MetaWorld trajectories into \emph{critical} and \emph{non-critical} phases.
    \textbf{Left}: representative trajectories show that truncation is triggered much more frequently during critical phases.
    \textbf{Right}: 83.7\% of truncations occur in critical phases, while only 16.3\% occur in non-critical phases.
    }
    \label{fig:lvm_phase_analysis}
\end{figure*}

LVM also intervenes at meaningful moments. We manually divide MetaWorld trajectories into \emph{critical} phases, such as precise grasping and alignment, and \emph{non-critical} phases, such as tolerant transport after a stable grasp. 
Figure~\ref{fig:lvm_phase_analysis} shows that 83.7\% of truncations occur in critical phases. This result verifies the core adaptive-horizon intuition of VLA-Corrector: it does not simply shorten the horizon everywhere, but instead preserves long-horizon efficiency in tolerant phases and restores short-horizon precision when the task enters error-sensitive phases.
\begin{figure}[t]
    \centering
    \includegraphics[width=0.95\columnwidth]{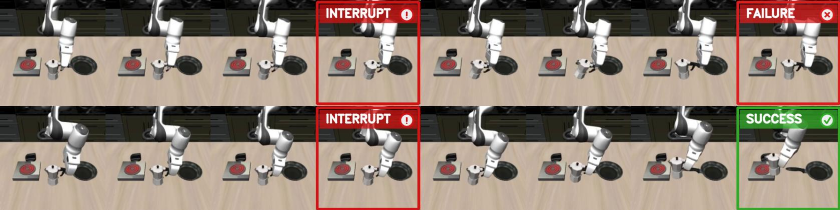}
    \caption{
    \textbf{Controlled recovery case.}
    Given the same initial state and detected grasping error, the monitored baseline continues the original chunk and fails (\textbf{top}), while VLA-Corrector truncates stale actions, replans with OGG, and completes the task (\textbf{bottom}).
    }
    \label{fig:qualitative_recovery}
\end{figure}

\textbf{OGG Correction.}
We isolate OGG by comparing standard re-inference and OGG-guided re-inference 
\begin{wrapfigure}{r}{0.48\columnwidth}
    \vspace{0pt}
    \centering
    \includegraphics[width=0.46\columnwidth]{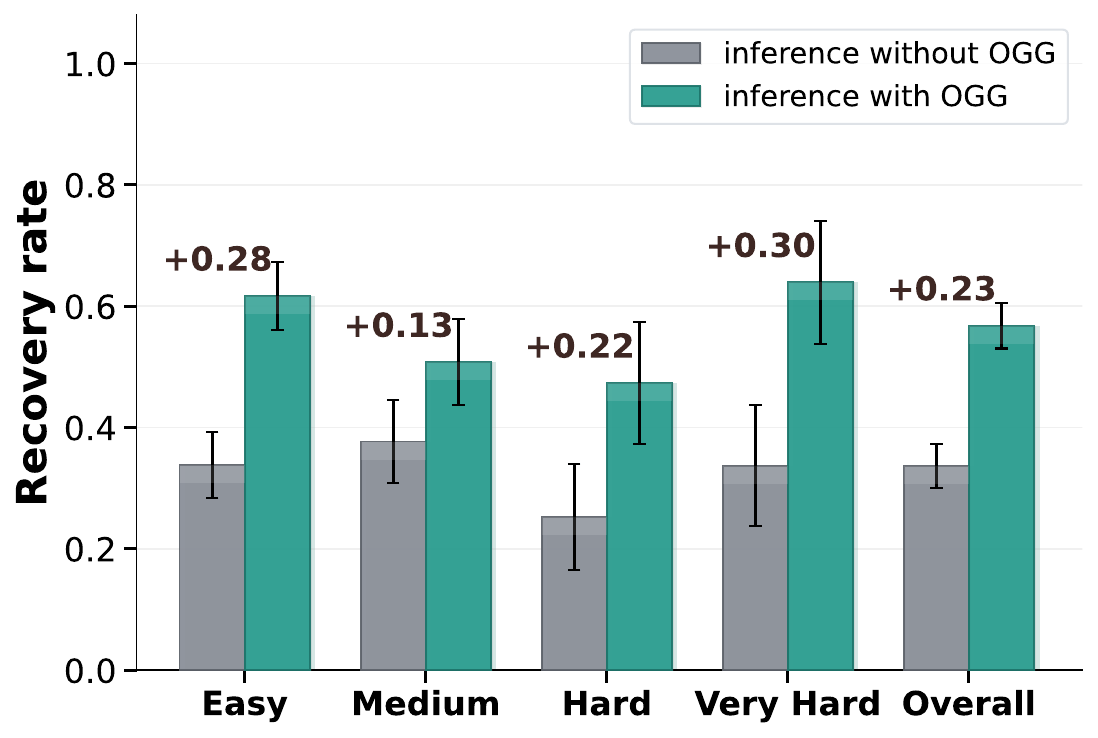}
    \caption{\textbf{Post-interrupt recovery.} OGG-guided inference consistently outperforms standard inference.}
    \label{fig:recovery_rate}
    \vspace{0pt}
\end{wrapfigure}
after the same interrupt-triggered truncation. Recovery is counted when $E_t<T_{\mathrm{off}}$ within the next 10 steps. As shown in Fig.~\ref{fig:recovery_rate}, OGG improves recovery across all difficulty levels, with an average gain of 0.23. This indicates that, after truncation stops stale actions, OGG-guided re-inference further improves the quality of the recovery replan compared to standard re-inference.


\textbf{Controlled Recovery Case.}
Figure~\ref{fig:qualitative_recovery} shows a representative LIBERO episode where both methods start from the same initial state and encounter the same grasping error when approaching the cup handle. 
In the top row, the baseline rollout is only monitored by LVM for error alignment without truncating the remaining actions; after the deviation is detected, it continues executing the original action chunk, leading to an unstable grasp, cup drop, and task failure. 
In the bottom row, VLA-Corrector detects the same deviation but actively truncates the stale actions and triggers corrective replanning, allowing the robot to recover a stable grasp and place the cup at the target location.
This controlled comparison highlights the overall effectiveness of VLA-Corrector under the same initial condition and the same detected execution error.

\subsection{Real-World Evaluation}

\textbf{Setup.}
We evaluate VLA-Corrector on an AgileX PiPER 6-DoF arm using $\pi_{0.5}$ as the backbone. The benchmark includes three task groups, each with three tasks and 20 trials: \textbf{1) Pick-and-place} tests standard manipulation, \textbf{2) Alignment} tests precision-sensitive placement or insertion, and \textbf{3) Disturbance recovery} tests recovery when objects or targets are manually shifted during execution. Full task definitions, protocols, failure analysis, and representative demonstrations are provided in Appendix~\ref{app:real_world_details}.

\begin{table}[hbtp]
  \centering
  \small
  \setlength{\tabcolsep}{4.2pt}
  \renewcommand{\arraystretch}{1.12}
  \caption{\textbf{Real-world evaluation on AgileX PiPER.}
  Success rate (\%, $\uparrow$) over three task groups. Each group contains three tasks with 20 trials per task. 
  We report 95\% binomial confidence intervals.}
  \label{tab:real_world_results}
  \begin{tabular}{lcccc}
    \toprule
    \rowcolor{headerbg}
    \textbf{Method} 
    & \textbf{Pick-place}$\uparrow$
    & \textbf{Alignment}$\uparrow$
    & \textbf{Disturbance}$\uparrow$
    & \textbf{Avg.}$\uparrow$ \\
    \midrule
    $\pi_{0.5}$ Baseline 
    & 70.0 $\pm$ 11.6
    & 56.7 $\pm$ 12.5
    & 40.0 $\pm$ 12.4
    & 55.6 $\pm$ 7.3 \\
    \rowcolor{softgreen}
    \textbf{+ VLA-Corrector} 
    & \textbf{78.3 $\pm$ 10.4} \gup{8.3}
    & \textbf{73.3 $\pm$ 11.2} \gup{16.6}
    & \textbf{68.3 $\pm$ 11.8} \gup{28.3}
    & \textbf{73.3 $\pm$ 6.5} \gup{17.7} \\
    \bottomrule
  \end{tabular}
\end{table}
\textbf{Results.}
Table~\ref{tab:real_world_results} shows that VLA-Corrector improves all three task groups, increasing average success from 55.6\% to 73.3\%. The gain is modest on pick-and-place (+8.3), larger on alignment (+16.6), and largest on disturbance recovery (+28.3). This trend matches the intended role of VLA-Corrector: it preserves standard execution performance, improves precision-sensitive manipulation, and is most beneficial when online disturbances make the remaining action chunk outdated.

\subsection{Ablation Studies}
\begin{table*}[htbp]
  \centering
  \small
  \setlength{\tabcolsep}{6pt}
  \renewcommand{\arraystretch}{1.12}
  \caption{\textbf{Component ablation on MetaWorld.}
  Success rate (\%, $\uparrow$). Green arrows indicate average improvement over the baseline.}
  \label{tab:ablation_components}
  \begin{tabular}{lccccc}
    \toprule
    \rowcolor{headerbg}
    \textbf{Variant} & \textbf{Easy}$\uparrow$ & \textbf{Medium}$\uparrow$ & \textbf{Hard}$\uparrow$ & \textbf{Very Hard}$\uparrow$ & \textbf{Avg.}$\uparrow$ \\
    \midrule
    Baseline (Open-loop) & 70.5 & 45.0 & 38.3 & 41.0 & 48.70 \\
    \rowcolor{softblue}
    + Truncation Only & 81.8 & 53.6 & \textbf{50.0} & 56.0 & 60.35 \gup{11.65} \\
    \rowcolor{softgreen}
    \textbf{+ Truncation + OGG} & \textbf{83.2} & \textbf{61.7} & 47.5 & \textbf{65.0} & \textbf{64.35} \gup{15.65} \\
    \bottomrule
  \end{tabular}
\end{table*}
We ablate VLA-Corrector on $\pi_{0.5}$ in MetaWorld, focusing on \textbf{(1) components}: necessity of truncation and OGG; \textbf{(2) detector design}: decoupling monitoring from backbone fine-tuning; and \textbf{(3) sensitivity}, with detailed results in Appendix~\ref{app:ablation_sensitivity}.

\textbf{Component effects.}
Table~\ref{tab:ablation_components} shows that truncation alone raises average success from 48.70\% to 60.35\%, confirming the importance of stopping stale actions. Adding OGG further improves the average to 64.35\%, showing that post-truncation recovery also benefits from guided re-inference.

\textbf{Detector design and sensitivity.}
We compare our decoupled LVM with an internal detection head trained during $\pi_{0.5}$ fine-tuning using the same auxiliary latent-residual prediction target. The internal head takes the final-layer hidden state of the last token, combines it with the executed action, and predicts the same task as the external LVM. As shown in Appendix~\ref{app:ablation_sensitivity}, decoupled LVM improves average success from 49.55\% to 64.35\%, suggesting that separating monitoring from action generation preserves the backbone representation and yields a cleaner drift signal. The same appendix also shows that moderate guidance strength and a 40M LVM are sufficient; overly strong guidance or larger monitors provide little additional benefit.

\begin{samepage}
\section{Conclusion}

This paper studies the open-loop blind spot in action-chunked VLA policies, where fixed horizons improve policy-call efficiency but allow stale actions to accumulate errors. VLA-Corrector addresses this issue with a lightweight detect-and-correct layer that monitors latent visual dynamics, truncates stale actions when drift persists, and guides the next inference toward recovery. Our results show that small inference-time modules can provide targeted robustness gains without retraining the VLA backbone. Rather than replacing action chunking, VLA-Corrector makes it adaptive: long-horizon execution is preserved when reliable, while corrective replanning is invoked when the current chunk should no longer be trusted.

\end{samepage}

\bibliographystyle{assets/acl_natbib}
\bibliography{paper}

\clearpage
\beginappendix
\section{Related Work}
\label{app:related_work}

\subsection{Generative VLA Models}
Embodied intelligence is shifting from task-specific policies toward unified Vision-Language-Action (VLA) foundation models. Representative works have shown that combining large-scale vision-language pretraining with robot control can substantially improve cross-task generalization, semantic understanding, and cross-platform transfer \cite{o2024open,black2024pi_0,zitkovich2023rt,kim2024openvla,ghosh2024octo,li2023vision,bjorck2025gr00t,pertsch2025fast,song2025accelerating}. On the action modeling side, modern VLA and general robot policies often adopt generative frameworks to model high-dimensional, continuous, and multi-modal action distributions: diffusion models generate actions through iterative denoising \cite{chi2025diffusion,wen2025diffusionvla,black2024pi_0,chen2025fast,li2025cogvla,zhao2025cot,hou2025dita,kim2025fine,song2025hume}, while flow matching directly learns velocity fields along continuous trajectories. These approaches typically offer stronger action expressivity, but they also incur higher policy-call cost\cite{wang2024one}. Consequently, reducing control latency without sacrificing generation quality has become a central challenge in deploying VLA models.

\subsection{Action Chunk and Horizon Trade-off}
To alleviate the tension between high-frequency control and expensive policy inference, \textbf{action chunk} has been widely adopted in continuous control. The core idea is to predict multiple future actions in a single inference call and then execute them sequentially with a downstream controller \cite{chi2025diffusion,zhao2023act,black2024pi_0,shukor2025smolvla,bjorck2025gr00t,wang2025sigma,driess2025knowledge}. This design effectively amortizes policy-call frequency and improves temporal smoothness, but it also reduces the use of fresh observations within a fixed action horizon. A longer horizon further lowers inference frequency and maintains action consistency, but weakens responsiveness to environmental changes; a shorter horizon more closely resembles strict closed-loop control and often provides better local recovery, yet requires much more frequent policy calls \cite{jing2025mixture,zhangaction,liu2024bidirectional,khan2025test,so2025improving,fu2025deep,zhao2026action,liu2026long}. Most existing approaches therefore seek a compromise under a static horizon, whereas our work focuses on an event-triggered dynamic horizon: the system retains the efficiency of long chunks during stable phases and truncates them on demand to recover closed-loop responsiveness during local failures.

\subsection{Failure Recovery in Visuomotor Policies}
Long-horizon visuomotor control commonly suffers from compounding errors and covariate shift: once execution drifts away from the training distribution, the policy often cannot recover on its own. Existing approaches can be broadly divided into two categories. One line of work expands the training distribution through additional interaction, human intervention, or recovery data collection, and then updates the policy accordingly \cite{sharma2023self,ingelhag2025real,hu2025rac,neary2025improving}. Another line explicitly introduces recovery policies, search mechanisms, or distributional constraints to pull the system back to a safe region once it departs from the expert manifold \cite{gao2025out,sun2025latent,jain2025smooth,liang2026adaptive,gou2024learning,wu2026speedup,xuvla}. These works collectively show that explicit recovery mechanisms can substantially improve the robustness of visuomotor policies; however, they typically rely on additional recovery data, separately trained recovery modules, or joint optimization tailored to a specific policy backbone.


\section{Method Details}
\label{app:method_details}

\subsection{Details of Event-Triggered Truncation}
\label{app:event_trigger_details}

This section provides the full event-triggered truncation rule used by LVM. Given the online inconsistency score $E_t$, the monitor maintains a sliding window of recent scores,
\[
\mathbf{E}_W=\{E_{t-w+1},\dots,E_t\}.
\]
Let
\[
M_e=\mathrm{median}(\mathbf{E}_W),
\qquad
\mathrm{MAD}=\mathrm{median}\!\left(|E_i-M_e|\right),\quad E_i\in\mathbf{E}_W .
\]
We use the median absolute deviation because it is less sensitive to transient score spikes than the mean and variance. Based on this robust statistic, we define asymmetric on/off thresholds,
\[
T_{\mathrm{on}}=M_e+\lambda_{\mathrm{on}}\mathrm{MAD},
\qquad
T_{\mathrm{off}}=M_e+\lambda_{\mathrm{off}}\mathrm{MAD},
\qquad
\lambda_{\mathrm{on}}>\lambda_{\mathrm{off}}.
\]
The higher threshold $T_{\mathrm{on}}$ is used to confirm abnormal deviation, while the lower threshold $T_{\mathrm{off}}$ provides hysteresis and prevents rapid oscillation between normal and abnormal states.

To avoid triggering on isolated spikes, we maintain a persistence counter:
\begin{equation}
c_t =
\begin{cases}
c_{t-1}+1, & \text{if } E_t>T_{\mathrm{on}},\\
0, & \text{if } E_t<T_{\mathrm{off}},\\
c_{t-1}, & \text{otherwise}.
\end{cases}
\label{eq:debounce_counter_app}
\end{equation}
An interrupt event is triggered when
\begin{equation}
c_t\ge p,
\label{eq:meltdown_trigger_app}
\end{equation}
where $p$ is the patience parameter. After an interrupt event, the remaining actions in the current execution queue are discarded, the counter is reset, and the next policy query is performed under corrective mode. If $h$ actions have already been executed from the current queue, the realized action horizon is
\[
H_{\mathrm{adaptive}}=h<H.
\]
In this way, the system preserves long-horizon execution during stable phases, while shortening the horizon when persistent visual drift indicates that the current action chunk is no longer reliable.

\subsection{LVM and OGG Runtime Parameters}
\label{app:runtime_parameters}

\textbf{LVM monitoring parameters.}
The Latent-space Vision Monitor uses a robust online thresholding state machine. We maintain a sliding window of recent inconsistency scores with window size 15. The dynamic thresholds are computed using $\lambda_{\mathrm{on}}=3.0$ and $\lambda_{\mathrm{off}}=2.0$. An interrupt event is triggered only when the inconsistency score exceeds the activation threshold for $p=5$ consecutive steps. Similarly, the monitor requires 5 consecutive safe steps for reset. To avoid repeated triggers immediately after intervention, we use an interrupt cooldown of 10 steps.

\textbf{OGG parameters.}
Online Gradient Guidance is invoked only for the single policy query immediately following an interrupt-triggered truncation. When an interrupt event occurs, the remaining action queue is cleared, the next replan is marked as a guidance replan, and OGG is applied during that next action-chunk generation. Subsequent policy calls return to standard inference unless another interrupt event is detected. The guidance strength is controlled by $\eta$. In our experiments, we use $\eta=1$ as the default setting for $\pi_{0.5}$, SmolVLA, and X-VLA. The sensitivity of performance to $\eta$ is analyzed in Table~\ref{tab:eta_ablation}.


\section{Training and Implementation Details}
\label{app:implementation_details}

\subsection{Benchmarks and Evaluation Protocol}
\label{app:benchmark_protocol}

We evaluate VLA-Corrector on three simulation benchmarks: MetaWorld~\cite{yu2020meta}, LIBERO~\cite{liu2023libero}, and RoboTwin v2.0. MetaWorld tests contact-rich manipulation robustness across difficulty splits, LIBERO evaluates language-conditioned long-horizon task execution, and RoboTwin broadens the evaluation to 50 tasks under clean and randomized scene configurations. We use $\pi_{0.5}$~\cite{intelligence2025pi_} as the main backbone, and further test SmolVLA~\cite{shukor2025smolvla} and X-VLA~\cite{zheng2025x} for cross-architecture evaluation. We report task success rate as the primary metric, together with policy calls, success-per-call efficiency, and post-interrupt recovery rate when applicable.

For MetaWorld, tasks are grouped into four difficulty splits: \emph{Easy}, \emph{Medium}, \emph{Hard}, and \emph{Very Hard}. All online evaluations are conducted task by task within each split, and we report the average success rate over tasks in the corresponding split. Unless otherwise specified, each task is evaluated with 20 episodes. For the corrector data-scaling study, we use a fixed episode-level split before subsampling the training data: 10\% of episodes are held out for validation, 10\% are held out for testing, and the remaining episodes form the training pool. We then subsample this training pool according to the training ratio $r$. In the multi-seed data-scaling analysis, we use split seeds $\{42,123,999\}$ and require at least 20 episodes in the train, validation, and test partitions. All simulation experiments are conducted on 8 NVIDIA A100-SXM4-40GB GPUs.

For RoboTwin v2.0, we evaluate all 50 tasks in both the clean and randomized configurations, using 50 episodes per task and configuration. Initial conditions and evaluation seeds are paired across methods, yielding 5,000 episodes per method. We compare the original fixed-horizon baseline ($H=50$), a latency-matched baseline ($H=30$), and VLA-Corrector ($H=50$), all with the same frozen $\pi_{0.5}$ backbone. We report clean, randomized, and pooled success rates with 95\% Wilson confidence intervals, together with average policy calls per episode and measured wall-clock latency per environment step.

\subsection{External Corrector Architecture}
\label{app:corrector_architecture}

The external corrector $M_\phi$ is implemented as a residual MLP for short-horizon latent dynamics prediction. We first map the executed action through a linear embedding layer, concatenate the resulting action feature with the current visual latent $Z_t^{\mathrm{real}}$, and feed the combined representation into a stack of residual MLP blocks. The network predicts the short-horizon latent residual rather than the absolute future latent state. In all experiments, we use a four-layer hidden configuration with width $[2048,2048,2048,2048]$. The exact number of parameters varies slightly across benchmarks and VLA backbones because both the action dimensionality and the visual latent dimensionality differ across settings. Overall, the correctors used in our experiments contain approximately 38--42M parameters, and we refer to this external module as a lightweight $\sim$40M MLP corrector.

\subsection{Corrector Training}
\label{app:corrector_training}

The corrector is trained after the VLA backbone has been fine-tuned on the benchmark training set. We freeze the VLA backbone and use its visual encoder to extract latent representations from demonstration trajectories. The corrector is then trained on these frozen latents to predict the short-horizon visual latent residual induced by the executed action. We use AdamW as the optimizer with learning rate $3\times10^{-4}$ and weight decay $10^{-4}$. The learning rate is scheduled by cosine annealing with $\eta_{\min}=0.01\times\mathrm{lr}$. We train for 30 epochs with early stopping patience 5. For the ratio-sweep training runs, we use batch size 128 and evaluation batch size 256. For the deployed h1-k10 correctors used in evaluation, we use batch size 512, train for 30 epochs, and optimize a cosine-based training loss. The number of optimization steps depends on the number of training samples under each data ratio:
\[
\mathrm{steps\ per\ epoch}=\left\lceil \frac{N_{\mathrm{train}}}{B} \right\rceil,
\qquad
\mathrm{total\ steps}=30\times \mathrm{steps\ per\ epoch},
\]
where $N_{\mathrm{train}}$ is the number of training samples and $B$ is the batch size.

\subsection{Compute Resources}
\label{app:compute_resources}

All simulation experiments are conducted on NVIDIA A100-SXM4-40GB GPUs. The main training and evaluation jobs are run on a server with 8 such GPUs. Corrector training is lightweight compared with VLA backbone fine-tuning because it only optimizes the external MLP on frozen VLA features. During online evaluation, LVM monitoring adds only a forward pass through the external corrector, while OGG introduces additional gradient computation only for the single recovery query following an interrupt event.


\section{Additional Experimental Results}
\label{app:additional_experiments}

\subsection{Ablation Sensitivity}
\label{app:ablation_sensitivity}

We provide additional ablations on detector design, OGG guidance strength, and LVM capacity. All experiments are conducted with $\pi_{0.5}$ on MetaWorld.

\begin{table*}[t]
  \centering
  \small
  \setlength{\tabcolsep}{5pt}
  \renewcommand{\arraystretch}{1.12}
  \caption{\textbf{Coupled vs. decoupled detection.}
  Success rate (\%, $\uparrow$). Both variants use OGG; only the detector design differs.}
  \label{tab:detection_architecture}
  \begin{tabular}{lccccc}
    \toprule
    \rowcolor{headerbg}
    \textbf{Detector} & \textbf{Easy}$\uparrow$ & \textbf{Medium}$\uparrow$ & \textbf{Hard}$\uparrow$ & \textbf{Very Hard}$\uparrow$ & \textbf{Avg.}$\uparrow$ \\
    \midrule
    Internal Head + OGG & 65.2 & 46.0 & 36.5 & 50.5 & 49.55 \\
    \rowcolor{oursbg}
    \textbf{Decoupled LVM + OGG}
      & \textbf{83.2} \gain{18.0}
      & \textbf{61.7} \gain{15.7}
      & \textbf{47.5} \gain{11.0}
      & \textbf{65.0} \gain{14.5}
      & \textbf{64.35} \gain{14.80} \\
    \bottomrule
  \end{tabular}
\end{table*}

For the internal detector, we fine-tune $\pi_{0.5}$ with an additional auxiliary loss that matches the prediction task of LVM: the final-layer hidden state of the last token is combined with the executed action and used to predict the same short-horizon latent residual. Table~\ref{tab:detection_architecture} shows that this coupled design performs much worse than the decoupled external LVM. This suggests that joint auxiliary detection training can interfere with the action-generation representation, whereas the external LVM preserves the backbone policy and provides a cleaner monitoring signal.

\begin{table*}[t]
  \centering
  \small
  \setlength{\tabcolsep}{7pt}
  \renewcommand{\arraystretch}{1.12}
  \caption{
  \textbf{Ablation on OGG guidance strength $\eta$.}
  Success rate (\%, $\uparrow$) on MetaWorld. The highlighted row is the default setting.
  }
  \label{tab:eta_ablation}
  \begin{tabular}{lccccc}
    \toprule
    \rowcolor{headerbg}
    \textbf{Guidance Strength} & \textbf{Easy}$\uparrow$ & \textbf{Medium}$\uparrow$ & \textbf{Hard}$\uparrow$ & \textbf{Very Hard}$\uparrow$ & \textbf{Avg.}$\uparrow$ \\
    \midrule
    $\eta = 0.1$
      & 82.7
      & 56.8
      & 45.2
      & \textbf{66.0}
      & 62.68 \\
    \rowcolor{softgreen}
    $\eta = 1$ \textbf{(default)}
      & \textbf{83.2} \gup{0.5}
      & \textbf{61.7} \gup{4.9}
      & \textbf{47.5} \gup{2.3}
      & 65.0 \rdown{1.0}
      & \textbf{64.35} \gup{1.67} \\
    $\eta = 10$
      & 82.5
      & 56.8
      & 38.3
      & 65.0
      & 60.65 \rdown{3.70} \\
    $\eta = 100$
      & 80.9
      & 55.0
      & 36.7
      & 63.0
      & 58.90 \rdown{5.45} \\
    \bottomrule
  \end{tabular}
\end{table*}

\begin{table*}[t]
  \centering
  \small
  \setlength{\tabcolsep}{7pt}
  \renewcommand{\arraystretch}{1.12}
  \caption{
  \textbf{Ablation on LVM capacity.}
  Success rate (\%, $\uparrow$) on MetaWorld. The highlighted row is the default setting.
  }
  \label{tab:monitor_size_ablation}
  \begin{tabular}{lccccc}
    \toprule
    \rowcolor{headerbg}
    \textbf{Monitor Capacity} & \textbf{Easy}$\uparrow$ & \textbf{Medium}$\uparrow$ & \textbf{Hard}$\uparrow$ & \textbf{Very Hard}$\uparrow$ & \textbf{Avg.}$\uparrow$ \\
    \midrule
    LVM-10M
      & 78.5
      & 52.4
      & 39.8
      & 55.6
      & 56.58 \\
    \rowcolor{softgreen}
    \textbf{LVM-40M (default)}
      & \textbf{83.2} \gup{4.7}
      & 61.7 \gup{9.3}
      & 47.5 \gup{7.7}
      & \textbf{65.0} \gup{9.4}
      & \textbf{64.35} \gup{7.77} \\
    LVM-160M
      & 82.8 \rdown{0.4}
      & \textbf{62.1} \gup{0.4}
      & \textbf{48.0} \gup{0.5}
      & 64.2 \rdown{0.8}
      & 64.28 \rdown{0.07} \\
    \bottomrule
  \end{tabular}
\end{table*}

Tables~\ref{tab:eta_ablation} and~\ref{tab:monitor_size_ablation} show that VLA-Corrector does not require overly strong guidance or excessively large monitors. $\eta=1$ achieves the best average success, while larger guidance weakens performance on harder tasks. Increasing LVM capacity from 10M to 40M substantially improves success, but 160M provides almost no additional average gain.

\subsection{Cross-Domain Corrector Generalization}
\label{app:cross_domain_corrector}

To examine whether the external corrector learns a transferable latent-dynamics signal rather than only memorizing benchmark-specific trajectories, we evaluate cross-domain corrector transfer on MetaWorld. We use the same $\pi_{0.5}$ MetaWorld baseline policy and compare two VLA-Corrector variants: one using a corrector trained on LIBERO demonstrations, and the other using a corrector trained on MetaWorld demonstrations. Both are evaluated on MetaWorld under the same protocol.

\begin{table*}[t]
  \centering
  \small
  \setlength{\tabcolsep}{4.8pt}
  \renewcommand{\arraystretch}{1.12}
  \caption{\textbf{Cross-domain corrector generalization on MetaWorld.}
  Success rate (\%, $\uparrow$) of the same $\pi_{0.5}$ MetaWorld baseline equipped with correctors trained from different demonstration domains. Green arrows indicate absolute improvement over the baseline.}
  \label{tab:cross_domain_corrector}
  \begin{tabular}{lccccc}
    \toprule
    \rowcolor{headerbg}
    \textbf{Method} 
    & \textbf{Easy}$\uparrow$ 
    & \textbf{Medium}$\uparrow$ 
    & \textbf{Hard}$\uparrow$ 
    & \textbf{Very Hard}$\uparrow$ 
    & \textbf{Avg.}$\uparrow$ \\
    \midrule
    $\pi_{0.5}$ Baseline 
      & 70.5 
      & 45.0 
      & 38.3 
      & 41.0 
      & 48.7 \\
    + LIBERO-trained Corrector
      & 70.9 \gup{0.4}
      & 50.5 \gup{5.5}
      & 40.7 \gup{2.4}
      & 45.0 \gup{4.0}
      & 51.8 \gup{3.1} \\
    \rowcolor{oursbg}
    + MetaWorld-trained Corrector
      & \textbf{74.1} \gup{3.6}
      & \textbf{53.8} \gup{8.8}
      & \textbf{52.9} \gup{14.6}
      & \textbf{54.0} \gup{13.0}
      & \textbf{58.7} \gup{10.0} \\
    \bottomrule
  \end{tabular}
\end{table*}

Table~\ref{tab:cross_domain_corrector} shows that the LIBERO-trained corrector still improves the MetaWorld baseline from 48.7\% to 51.8\%, suggesting that the learned latent-dynamics consistency signal has limited but non-trivial cross-domain transfer. However, the MetaWorld-trained corrector achieves a much larger gain, improving the average success rate to 58.7\%. This indicates that while the corrector can capture partially transferable notions of on-track visual dynamics, domain-matched demonstrations remain important for accurate deviation detection and effective corrective guidance.

\subsection{Full Performance--Efficiency Results}
\label{app:full_efficiency}

Table~\ref{tab:comprehensive_efficiency} provides the full performance--efficiency results across all evaluated VLA backbones and action horizons. These results complement the main-text analysis by reporting the exact success rates, average policy calls, and success-per-call efficiency gains for each setting. Overall, VLA-Corrector consistently improves task success while maintaining or reducing the number of policy calls in most cases, leading to positive success-per-call efficiency gains across all backbones. The gains are especially pronounced at longer horizons, where the open-loop blind spot is larger and stale actions are more likely to accumulate errors before the next policy query.

\begin{table*}[t]
\centering
\small
\setlength{\tabcolsep}{3.2pt} 
\renewcommand{\arraystretch}{1.15}
\begin{threeparttable}
\caption{\textbf{Performance--efficiency trade-off across VLA backbones and action horizons.}
Success rate, average policy calls per episode, and relative efficiency gains are reported. 
The arrows ($\uparrow$/$\downarrow$) in the header explicitly indicate whether a higher or lower metric is optimal.}
\label{tab:comprehensive_efficiency}
\begin{tabular}{l c cc cc ccc}
\toprule
\multirow{2}{*}{\textbf{Model}} & \multirow{2}{*}{\textbf{Horizon}} & \multicolumn{2}{c}{\textbf{Baseline}} & \multicolumn{2}{c}{\textbf{Ours (LVM+OGG)}} & \multicolumn{3}{c}{\textbf{Improvements}} \\
\cmidrule(lr){3-4} \cmidrule(lr){5-6} \cmidrule(lr){7-9}
& & \textbf{Succ} (\%)$\uparrow$ & \textbf{Calls} $\downarrow$ & \textbf{Succ} (\%)$\uparrow$ & \textbf{Calls} $\downarrow$ & \textbf{$\Delta$Succ} (\%)$\uparrow$ & \textbf{$\Delta$Calls} $\downarrow$ & \textbf{Eff. Gain} (\%)$\uparrow$ \\
\midrule
\multirow{5}{*}{\textbf{$\pi_{0.5}$}}
& 10 & 64.50 & 20.41 & \cellcolor{softgreen}\textbf{72.40} & \cellcolor{softgreen}\textbf{17.64} & \gup{7.90} & \gdown{2.77} & \cellcolor{softblue}\textbf{+29.9\%} \\
& 20 & 61.40 & 10.77 & \cellcolor{softgreen}\textbf{69.10} & \cellcolor{softgreen}\textbf{9.95}  & \gup{7.70} & \gdown{0.82} & \cellcolor{softblue}\textbf{+21.8\%} \\
& 30 & 56.10 & 7.97  & \cellcolor{softgreen}\textbf{63.50} & \cellcolor{softgreen}\textbf{7.92}  & \gup{7.40} & \gdown{0.05} & \cellcolor{softblue}\textbf{+13.9\%} \\
& 40 & 54.45 & 5.96  & \cellcolor{softgreen}\textbf{60.80} & \cellcolor{softgreen}6.04 & \gup{6.35} & \rup{0.08} & \cellcolor{softblue}\textbf{+10.2\%} \\
& 50 & 48.72 & 5.15  & \cellcolor{softgreen}\textbf{58.70} & \cellcolor{softgreen}\textbf{4.98} & \gup{9.98} & \gdown{0.17} & \cellcolor{softblue}\textbf{+24.6\%} \\
\midrule
\multirow{5}{*}{\textbf{SmolVLA}}
& 10 & 61.90 & 19.27 & \cellcolor{softgreen}\textbf{73.00} & \cellcolor{softgreen}\textbf{15.64} & \gup{11.10} & \gdown{3.63} & \cellcolor{softblue}\textbf{+45.3\%} \\
& 20 & 58.90 & 11.32 & \cellcolor{softgreen}\textbf{68.90} & \cellcolor{softgreen}\textbf{9.50}  & \gup{10.00} & \gdown{1.82} & \cellcolor{softblue}\textbf{+39.4\%} \\
& 30 & 58.10 & 7.90  & \cellcolor{softgreen}\textbf{65.20} & \cellcolor{softgreen}\textbf{7.60}  & \gup{7.10} & \gdown{0.30} & \cellcolor{softblue}\textbf{+16.6\%} \\
& 40 & 56.80 & 5.64  & \cellcolor{softgreen}\textbf{67.20} & \cellcolor{softgreen}\textbf{5.13}  & \gup{10.40} & \gdown{0.51} & \cellcolor{softblue}\textbf{+30.0\%} \\
& 50 & 54.20 & 4.86  & \cellcolor{softgreen}\textbf{62.90} & \cellcolor{softgreen}\textbf{4.68}  & \gup{8.70} & \gdown{0.18} & \cellcolor{softblue}\textbf{+20.6\%} \\
\midrule
\multirow{5}{*}{\textbf{X-VLA}}
& 4  & 68.50 & 46.58 & \cellcolor{softgreen}\textbf{72.00} & \cellcolor{softgreen}\textbf{35.20} & \gup{3.50} & \gdown{11.38} & \cellcolor{softblue}\textbf{+39.1\%} \\
& 8  & 67.80 & 23.18 & \cellcolor{softgreen}\textbf{72.10} & \cellcolor{softgreen}\textbf{19.48} & \gup{4.30} & \gdown{3.70} & \cellcolor{softblue}\textbf{+26.5\%} \\
& 16 & 55.50 & 13.71 & \cellcolor{softgreen}\textbf{60.90} & \cellcolor{softgreen}13.92 & \gup{5.40} & \rup{0.21} & \cellcolor{softblue}\textbf{+8.1\%} \\
& 24 & 52.90 & 10.13 & \cellcolor{softgreen}\textbf{59.20} & \cellcolor{softgreen}\textbf{9.92} & \gup{6.30} & \gdown{0.21} & \cellcolor{softblue}\textbf{+14.3\%} \\
& 32 & 44.00 & 8.61  & \cellcolor{softgreen}\textbf{54.40} & \cellcolor{softgreen}\textbf{8.34} & \gup{10.40} & \gdown{0.27} & \cellcolor{softblue}\textbf{+27.6\%} \\
\bottomrule
\end{tabular}
\begin{tablenotes}[flushleft]
\footnotesize
\item \textbf{$\Delta$Succ}: absolute success-rate improvement. 
\textbf{$\Delta$Calls}: absolute change in policy calls per episode. 
\textbf{Eff. Gain}: relative improvement in success-per-call over the baseline.
\end{tablenotes}
\end{threeparttable}
\end{table*}

\subsection{Data Efficiency of Corrector Training}
\label{app:data_efficiency}

\begin{table*}[htbp]
  \centering
  \small
  \caption{\textbf{Data efficiency of corrector training on MetaWorld.}
  We compare the original open-loop baseline at NAS=50 with VLA-Corrector trained using different fractions of demonstration trajectories. The results show that moderate corrector data already surpasses the baseline and that performance largely saturates around 60\% of the training set. Bold indicates the best result in each column.}
  \label{tab:data_efficiency}
  \begin{tabular}{lccccc}
    \toprule
    \textbf{Method / Training Ratio} & \textbf{Easy} & \textbf{Medium} & \textbf{Hard} & \textbf{Very Hard} & \textbf{Macro Avg.} \\
    \midrule
    Baseline (NAS=50) & 70.54 & 45.00 & 38.33 & 41.00 & 48.72 \\
    \midrule
    VLA-Corrector ($r = 0.2$) & 71.07 & 49.55 & 36.67 & 36.00 & 48.32 \\
    VLA-Corrector ($r = 0.4$) & 70.36 & 45.91 & 38.33 & 42.00 & 49.15 \\
    VLA-Corrector ($r = 0.6$) & 70.89 & 52.73 & 39.17 & \textbf{46.00} & 52.20 \\
    VLA-Corrector ($r = 0.8$) & 71.07 & 53.64 & 38.33 & \textbf{46.00} & 52.26 \\
    VLA-Corrector ($r = 1.0$) & \textbf{73.21} & \textbf{55.91} & \textbf{44.17} & 44.00 & \textbf{54.32} \\
    \bottomrule
  \end{tabular}
\end{table*}

\textbf{Analysis.}
To quantify how much data is needed to train the external corrector, we subsample the MetaWorld demonstration set with training ratios $r \in \{0.2,0.4,0.6,0.8,1.0\}$ and evaluate the resulting VLA-Corrector on the same task suite. As shown in Table~\ref{tab:data_efficiency}, the corrector is already effective with limited data: using only 20\% of the demonstrations yields a 48.32\% macro-average success rate. Performance improves as more data is added and reaches 52.20\% at $r=0.6$, after which the gain becomes marginal. This suggests that the corrector does not require exhaustive coverage of all task variations; once it has seen enough examples of on-track local dynamics, it can provide a useful signal for detecting deviations and guiding recovery.

The trend also shows that the corrector benefits from additional data, but with diminishing returns. Increasing the training ratio from $0.2$ to $0.6$ brings a clear improvement, while increasing it further from $0.6$ to $0.8$ changes the macro average only slightly. Full-data training still achieves the best overall result, especially on the Hard split, but the relatively strong performance at moderate data ratios supports our design motivation: VLA-Corrector learns a local consistency signal rather than a full policy or heavy world model, making it data-efficient to train. On the Very Hard split, the small fluctuation across higher ratios indicates that extremely challenging contact-rich tasks are influenced by factors beyond corrector data scale alone, such as disturbance severity and backbone recovery capability.

\subsection{Inference-Time Cost and Control Delay}
\label{app:inference_time}

The monitor runs in parallel with the robot's current action. We therefore report two different times: (1) the time needed to complete the monitor's computation, and (2) the part of that time that actually makes the controller wait. We also measure the extra time used by the OGG recovery call. All measurements use synchronized CUDA timing. The LVM measurements are listed in Table~\ref{tab:lvm_component_timing}, the OGG measurements in Table~\ref{tab:ogg_deployment_timing}, and the physical-control check in Table~\ref{tab:piper_control_timing}.

\begin{table*}[t]
  \centering
  \small
  \setlength{\tabcolsep}{8pt}
  \renewcommand{\arraystretch}{1.12}
  \caption{\textbf{LVM timing during deployment on MetaWorld.} We measure 500 held-out observations after removing 20 warm-up observations. ``Full LVM computation'' is all work done by the monitor; ``wait added to control loop'' is the part that the controller actually waits for.}
  \label{tab:lvm_component_timing}
  \begin{tabular}{lcc}
    \toprule
    \rowcolor{headerbg}
    \textbf{Component} & \textbf{Mean (ms)} & \textbf{p95 (ms)} \\
    \midrule
    Preprocessing & 1.79 & 2.06 \\
    Frozen vision encoder & 21.31 & 21.46 \\
    40M residual MLP & 2.79 & 2.93 \\
    Mismatch + threshold & 0.91 & 1.08 \\
    \rowcolor{softgreen}
    \textbf{Full LVM computation} & \textbf{26.96} & \textbf{27.57} \\
    \rowcolor{softblue}
    \textbf{Wait added to control loop} & \textbf{0.008} & \textbf{0.009} \\
    \bottomrule
  \end{tabular}
\end{table*}

As shown in Table~\ref{tab:lvm_component_timing}, the monitor performs $26.96$~ms of computation on average and $27.57$~ms at p95. Because this work runs while the robot is executing, it adds only $0.008$~ms of average waiting time ($0.009$~ms at p95) to the control loop. Thus, the computation is real, but almost all of it is hidden by parallel execution; the $0.008$~ms number is only the extra wait, not the total monitor cost.

\begin{table*}[t]
  \centering
  \small
  \setlength{\tabcolsep}{10pt}
  \renewcommand{\arraystretch}{1.12}
  \caption{\textbf{OGG timing during deployment on an NVIDIA A100-SXM4-40GB.} We compare 200 matched ordinary/OGG pairs after 20 warm-up runs. The random state is restored for each pair, the order is balanced, and CUDA timing is synchronized; every OGG call uses two guidance steps.}
  \label{tab:ogg_deployment_timing}
  \begin{tabular}{lc}
    \toprule
    \rowcolor{headerbg}
    \textbf{Call} & \textbf{Mean latency} \\
    \midrule
    Ordinary & 232.490~ms \\
    OGG & 243.223~ms \\
    \rowcolor{softgreen}
    \textbf{Increment per trigger} & \textbf{10.733~ms} \\
    95\% CI & [10.430, 11.053]~ms \\
    \bottomrule
  \end{tabular}
\end{table*}

Table~\ref{tab:ogg_deployment_timing} shows that an ordinary policy call takes $232.490$~ms, while an OGG-guided call takes $243.223$~ms. OGG therefore adds $10.733$~ms only when it is used for recovery after an interrupt; it does not slow down ordinary policy calls.

\begin{table*}[t]
  \centering
  \small
  \setlength{\tabcolsep}{10pt}
  \renewcommand{\arraystretch}{1.12}
  \caption{\textbf{Timing check on AgileX PiPER.} PiPER executes 10 actions per second, so the controller has 100~ms for each action.}
  \label{tab:piper_control_timing}
  \begin{tabular}{lc}
    \toprule
    \rowcolor{headerbg}
    \textbf{Quantity} & \textbf{Result} \\
    \midrule
    Control period & 100~ms/action (10~Hz) \\
    Full LVM & 26.96~ms mean / 27.57~ms p95 \\
    p95 remaining margin & 72.43~ms \\
    Deadline misses & 0 / 500 \\
    Wait added to control loop & 0.008~ms mean \\
    \bottomrule
  \end{tabular}
\end{table*}

\textbf{Discussion.}
Table~\ref{tab:piper_control_timing} summarizes the physical check. MetaWorld is evaluated without a real-time clock, but PiPER must complete one action every 100~ms. The monitor's p95 computation time is 27.57~ms, leaving 72.43~ms in each control period, and none of the 500 measured steps missed its deadline. The monitor therefore remains responsive in the physical setup: its full computation is accounted for, while only 0.008~ms of average waiting is added to the controller. OGG adds 10.733~ms only on the occasional recovery call.

\subsection{Asynchronous and RTC Comparisons}
\label{app:async_rtc_matched}

Because Real-Time Chunking (RTC) periodically refreshes action chunks while the previous chunk is executing, it overlaps with the asynchronous execution path of VLA-Corrector and is a natural comparison point. The two mechanisms nevertheless address different parts of the control problem: RTC schedules the next policy query, whereas VLA-Corrector uses the observed-vs.~predicted visual-dynamics mismatch to produce an explicit error signal, stop stale actions, and guide the recovery replan. We therefore report both the deployment-path timing and a direct overlapped-execution success comparison. Table~\ref{tab:async_summary} collects the asynchronous timing measurements, Table~\ref{tab:rtc_context} gives the RTC timing context, and Table~\ref{tab:overlapped_rtc_metaworld} reports the direct overlapped-execution experiment.

\begin{table*}[t]
  \centering
  \small
  \setlength{\tabcolsep}{7pt}
  \renewcommand{\arraystretch}{1.12}
  \caption{\textbf{Asynchronous deployment-path accounting.} Intrinsic computation is measured with CUDA synchronization; exposed wait is the delay visible to the control loop after asynchronous pipelining. OGG is applied only to the single recovery query following an interrupt.}
  \label{tab:async_summary}
  \begin{tabular}{lc}
    \toprule
    \rowcolor{headerbg}
    \textbf{Quantity} & \textbf{Measured result} \\
    \midrule
    Full LVM intrinsic computation & 26.96~ms mean / 27.57~ms p95 \\
    Exposed asynchronous control-path wait & 0.008~ms mean / 0.009~ms p95 \\
    Ordinary policy call & 232.490~ms \\
    OGG-guided policy call & 243.223~ms \\
    Increment on the recovery call & 10.733~ms/trigger, 95\% CI [10.430, 11.053]~ms \\
    PiPER control period & 100~ms/action (10~Hz) \\
    Observed deadline misses & 0/500 \\
    \bottomrule
  \end{tabular}
\end{table*}

Table~\ref{tab:async_summary} makes the deployment claim precise: asynchronous execution does not remove the LVM's 26.96~ms of intrinsic computation, but only 0.008~ms of mean waiting time is exposed on the control path. The OGG increment is similarly localized to the recovery call, rather than being charged to every action step.

\begin{table*}[t]
  \centering
  \scriptsize
  \setlength{\tabcolsep}{5pt}
  \renewcommand{\arraystretch}{1.12}
  \caption{\textbf{RTC context under the 10~Hz control-period model.} The response intervals are obtained from the ordinary/guided call latencies and the persistence rule ($p=5$); they clarify the difference between periodic scheduling and failure-aware truncation.}
  \label{tab:rtc_context}
  \begin{adjustbox}{max width=\textwidth}
  \begin{tabular}{lcccc}
    \toprule
    \rowcolor{headerbg}
    \textbf{Configuration} & \textbf{Requery rule} & \textbf{Calls/step} & \textbf{Relative calls} & \textbf{Response under the timing model} \\
    \midrule
    LVM-gated truncation & $H=50$; extra replan after persistent mismatch & 0.0379 & 1.0$\times$ & Stale suffix stopped at $\approx$400~ms \\
    Reviewer-style RTC & Periodic, $s=3$ & 0.333 & 8.8$\times$ & Replacement at 297--597~ms \\
    Call-matched RTC & Periodic, $s\approx26.4$ & 0.0379 & 1.0$\times$ & Replacement at $\approx$297--2,937~ms \\
    \bottomrule
  \end{tabular}
  \end{adjustbox}
\end{table*}

Table~\ref{tab:rtc_context} gives an analytical timing context rather than a hardware benchmark. It shows that periodic RTC can deliver a newer ordinary chunk, but does not itself provide an explicit failure signal or a recovery-directed target. At the same call budget, RTC must distribute policy calls uniformly, whereas LVM-gated execution allocates the additional replan after an observed persistent mismatch. The direct experiment below tests whether this distinction remains visible in task success when all methods use overlapped execution.

For clarity, all main-text experiments and the previously reported VLA-Corrector results use \linebreak[4]$\texttt{rtc\_config=None}$: RTC itself is disabled in those experiments. This setting does not make LVM monitoring synchronous; the deployment-path measurements above still run the monitor asynchronously while the current action executes. To compare against RTC directly, we additionally run the experiment in Table~\ref{tab:overlapped_rtc_metaworld} with asynchronous inference enabled for every method and with an RTC baseline included only for this comparison. For context, the RTC analysis reports 97~ms for inference-time RTC versus 76~ms for vanilla $\pi_{0.5}$ on an RTX~4090, because prefix inpainting requires gradient computation; its training-time variant removes that inference operation only through a separately trained prefix-conditioned $\pi_{0.6}$ checkpoint. These context figures are kept separate from the direct overlapped-execution experiment and the A100 deployment-path measurements above.

\begin{table*}[t]
  \centering
  \scriptsize
  \setlength{\tabcolsep}{7pt}
  \renewcommand{\arraystretch}{1.12}
  \caption{\textbf{Direct overlapped-execution comparison on MetaWorld-50.} All methods use the same frozen $\pi_{0.5}$, horizon $H=50$, asynchronous execution path, and 50 paired episodes per task (2,500 episodes per method). The RTC baseline periodically alternates $s=26$. The three $E_t$-gated variants share the same detector and truncation settings and differ only in the triggered replan; the random direction is scaled to the mean OGG guidance norm.}
  \label{tab:overlapped_rtc_metaworld}
  \begin{adjustbox}{max width=\textwidth}
  \begin{tabular}{lccc}
    \toprule
    \rowcolor{headerbg}
    \textbf{Method} & \textbf{Triggered guidance} & \textbf{Success (95\% CI)} & \textbf{Calls/100 steps} \\
    \midrule
    RTC, $s=26$ & None & 61.3 [59.4, 63.2] & 3.85 \\
    $E_t$-gated replan & None & 62.4 [60.5, 64.3] & 4.08 \\
    $E_t$-gated random & Random & 61.9 [60.0, 63.8] & 4.12 \\
    \rowcolor{oursbg}
    \textbf{$E_t$-gated OGG} & \textbf{OGG} & \textbf{67.1 [65.3, 68.9]} & \textbf{3.88} \\
    \bottomrule
  \end{tabular}
  \end{adjustbox}
\end{table*}

As shown in Table~\ref{tab:overlapped_rtc_metaworld}, the additional experiment directly evaluates VLA-Corrector under overlapped execution rather than inferring its behavior from sequential timing. At comparable policy-call rates, $E_t$-gated OGG reaches 67.1\% success versus 61.3\% for periodic RTC, an improvement of 5.8 points (paired 95\% CI [+4.4, +7.2], McNemar $p<10^{-13}$). Thus, the advantage persists when policy inference is overlapped with action execution and is not explained by a higher call frequency. The matched-norm random control reaches 61.9\%, which is 5.2 points below $E_t$-gated OGG (95\% CI [+3.8, +6.6], $p<10^{-11}$), and does not improve over the ordinary $E_t$-gated replan (61.9\% vs.~62.4\%). These results isolate the contribution of the recovery direction: the gain comes from using the LVM-derived error signal to guide OGG toward a corrective target, rather than from arbitrary perturbation or simply requesting another policy chunk.


\section{Real-World Details}
\label{app:real_world_details}

\subsection{Robot Platform and Task Suite}
\label{app:real_world_platform_tasks}

\textbf{Robot platform.}
All real-world experiments are conducted on an \textbf{AgileX PiPER} 6-DoF robotic arm. We use $\pi_{0.5}$ as the action-chunked VLA baseline and evaluate VLA-Corrector on top of the same fine-tuned backbone. Both methods share the same RGB camera observation, language instruction, action horizon, and initial task conditions. Each task is evaluated with 20 trials per method. A trial is counted as successful if the robot completes the specified task without human reset after execution starts.

\textbf{Task design.}
We design three groups of real-world tasks with increasing difficulty. The first group evaluates standard pick-and-place behavior, where the original action chunk often remains valid. The second group evaluates precise alignment, where small accumulated errors can cause missed placement, collision, or failed insertion. The third group introduces online human disturbances during precision-sensitive phases, directly testing whether VLA-Corrector can recover when the remaining action chunk becomes stale. Table~\ref{tab:real_world_task_details} summarizes the task definitions.

\begin{table*}[t]
  \centering
  \small
  \setlength{\tabcolsep}{5.5pt}
  \renewcommand{\arraystretch}{1.15}
  \caption{\textbf{Real-world task suite on AgileX PiPER.}
  The benchmark contains three task groups and nine tasks in total. The disturbance group reuses manipulation skills from the first two groups but introduces online pose changes during precision-sensitive phases.}
  \label{tab:real_world_task_details}
  \begin{tabular}{@{}lp{0.25\linewidth}p{0.48\linewidth}@{}}
    \toprule
    \rowcolor{headerbg}
    \textbf{Group} & \textbf{Task} & \textbf{Description} \\
    \midrule
    \multirow{3}{*}{Pick-and-place}
    & Cube-to-region 
    & Pick up a cube and place it inside a marked target region. \\
    & Cube-to-drawer-top 
    & Pick up a cube and place it on top of a small drawer platform. \\
    & Object-to-container 
    & Pick up a cube and place it into a specific bowl. \\
    \midrule
    \multirow{3}{*}{Alignment}
    & Corner placement 
    & Place a cube precisely at the upper-left corner of the drawer top. \\
    & Square-hole insertion 
    & Align and insert a small cube into a square hole or slot. \\
    & Edge alignment 
    & Align an object with a narrow marked edge or target strip. \\
    \midrule
    \multirow{3}{*}{Disturbance recovery}
    & Moving-object grasp 
    & Move the target object when the gripper is about to grasp it. \\
    & Moving-placement target 
    & Move the placement region after the object is grasped. \\
    & Moving-insertion target 
    & Move the square hole or alignment target before insertion. \\
    \bottomrule
  \end{tabular}
\end{table*}

\subsection{Per-Task Real-World Results}
\label{app:real_world_per_task_results}

Table~\ref{tab:real_world_per_task} reports per-task success rates. The gains are modest on pick-and-place tasks because the baseline can often complete the task when the object and target remain static. The gains become larger on alignment tasks, where small pose errors are less tolerable. The largest improvement appears in disturbance recovery tasks, where the current action chunk often becomes stale after the object or target is manually moved.

\begin{table*}[htbp]
  \centering
  \small
  \setlength{\tabcolsep}{5.5pt}
  \renewcommand{\arraystretch}{1.12}
  \caption{\textbf{Per-task real-world success rates.}
  Success rate (\%, $\uparrow$) over 20 trials per task on the AgileX PiPER 6-DoF arm.
  Group and overall rows report mean $\pm$ standard deviation across tasks. 
  Green arrows indicate absolute improvement over the $\pi_{0.5}$ baseline.}
  \label{tab:real_world_per_task}
  \begin{tabular}{llcc}
    \toprule
    \rowcolor{headerbg}
    \textbf{Group} & \textbf{Task} 
    & \textbf{$\pi_{0.5}$ Baseline}$\uparrow$ 
    & \textbf{+ VLA-Corrector}$\uparrow$ \\
    \midrule

    \multirow{4}{*}{Pick-and-place}
    & Cube-to-region        & 75.0 & \textbf{85.0} \gup{10.0} \\
    & Cube-to-drawer-top    & 70.0 & \textbf{80.0} \gup{10.0} \\
    & Object-to-container   & 65.0 & \textbf{70.0} \gup{5.0} \\
    \rowcolor{groupbg}
    & \textbf{Group Avg.}   & \textbf{70.0 $\pm$ 5.0} & \textbf{78.3 $\pm$ 7.6} \gup{8.3} \\
    \midrule

    \multirow{4}{*}{Alignment}
    & Corner placement      & 60.0 & \textbf{75.0} \gup{15.0} \\
    & Square-hole insertion & 50.0 & \textbf{70.0} \gup{20.0} \\
    & Edge alignment        & 60.0 & \textbf{75.0} \gup{15.0} \\
    \rowcolor{groupbg}
    & \textbf{Group Avg.}   & \textbf{56.7 $\pm$ 5.8} & \textbf{73.3 $\pm$ 2.9} \gup{16.6} \\
    \midrule

    \multirow{4}{*}{Disturbance recovery}
    & Moving-object grasp       & 45.0 & \textbf{75.0} \gup{30.0} \\
    & Moving-placement target   & 40.0 & \textbf{70.0} \gup{30.0} \\
    & Moving-insertion target   & 35.0 & \textbf{60.0} \gup{25.0} \\
    \rowcolor{groupbg}
    & \textbf{Group Avg.}       & \textbf{40.0 $\pm$ 5.0} & \textbf{68.3 $\pm$ 7.6} \gup{28.3} \\
    \midrule

    \rowcolor{softgreen}
    \textbf{Overall} & \textbf{Avg. across 9 tasks}
    & \textbf{55.6 $\pm$ 13.8}
    & \textbf{73.3 $\pm$ 7.1} \gup{17.7} \\
    \bottomrule
  \end{tabular}
\end{table*}

\subsection{Real-World Protocol and Task Details}
\label{app:real_world_protocol}

\textbf{Evaluation protocol.}
For each real-world task, we collect 10--30 human teleoperated demonstration trajectories and use them to fine-tune the $\pi_{0.5}$ backbone for that task. The same fine-tuned backbone is then used by both the baseline and VLA-Corrector, so the comparison isolates the effect of the inference-time correction module. For each task and each method, we run 20 evaluation trials. Initial object poses are randomized within a small predefined region while keeping the task feasible for the 6-DoF PiPER arm.

\textbf{Pick-and-place tasks.}
The pick-and-place group includes placing a cube into a marked region, placing a cube on top of a drawer platform, and placing a small object into a bowl. These tasks are relatively tolerant to small pose errors, so the original action chunk often remains usable. VLA-Corrector mainly helps when occasional drift occurs during approach or release.

\textbf{Alignment tasks.}
The alignment group increases the precision requirement. In corner placement, the robot must place a cube at a specific corner of the drawer top. In square-hole insertion, the cube must be aligned with a square opening before insertion. In edge alignment, the robot must align an object with a narrow visual marker or edge. These tasks are more sensitive to accumulated error, making stale action execution more harmful near the final contact.

\textbf{Disturbance recovery tasks.}
The disturbance group directly evaluates the open-loop blind spot. We introduce manual perturbations at predefined task phases rather than at fixed time steps. For moving-object grasp, the target object is shifted when the gripper is close to grasping it. For moving-placement target, the robot first grasps the object normally, and the target platform or placement region is shifted before release. For moving-insertion target, the square hole or alignment target is shifted when the robot is about to align or insert. The disturbance is applied once per trial, kept within the camera view, and constrained to a physically recoverable range. The same timing rule and disturbance range are used for both the baseline and VLA-Corrector.

\subsection{Observed Real-World Failure Modes}
\label{app:real_world_failure_modes}

Despite the improvements, VLA-Corrector still fails in several real-world cases. If the target object or fixture is moved beyond the reachable region, or if the disturbance happens after the gripper has entered an unfavorable pose, the robot may no longer have enough workspace or time to recover through a single corrective replan. In tight alignment tasks, a 6-DoF arm without force feedback may fail due to contact geometry, friction, or small height errors even when the visual target is corrected. We also observe failures caused by visual ambiguity, such as partial occlusion by the gripper or poor contrast between the object and the target region. Finally, OGG still depends on the frozen $\pi_{0.5}$ action prior: it can bias the next generation toward a better corrective direction, but it cannot create a recovery behavior that the backbone policy cannot represent.

\subsection{Real-World Demonstration Examples}
\label{app:real_world_demos}

We provide representative real-world demonstration examples mainly for the disturbance recovery tasks. In these cases, the object or target fixture is manually shifted during execution, allowing us to visualize the practical robustness of VLA-Corrector under human-induced disturbances.

\begin{figure*}[htbp]
    \centering
    \includegraphics[width=0.95\textwidth]{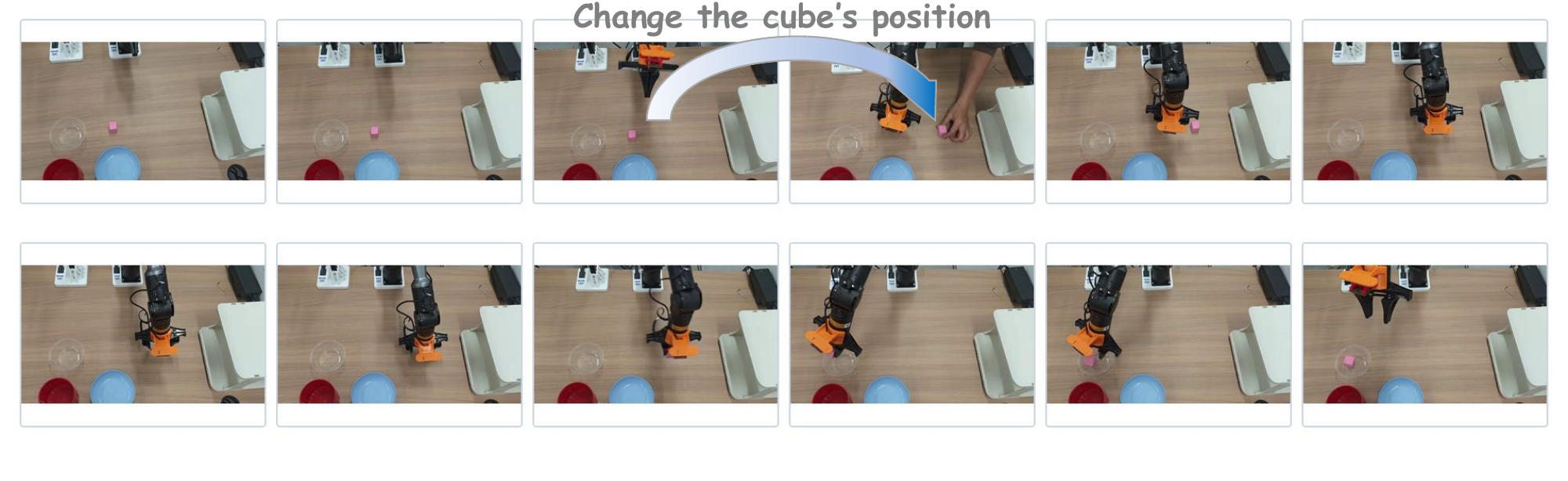}
    \caption{\textbf{Demo: Moving-object grasp.} The robot picks up the cube and places it into the white bowl, while a human manually changes the cube's position during the process.}
    \label{fig:demo_moving_object_grasp}
\end{figure*}

\begin{figure*}[htbp]
    \centering
    \includegraphics[width=0.95\textwidth]{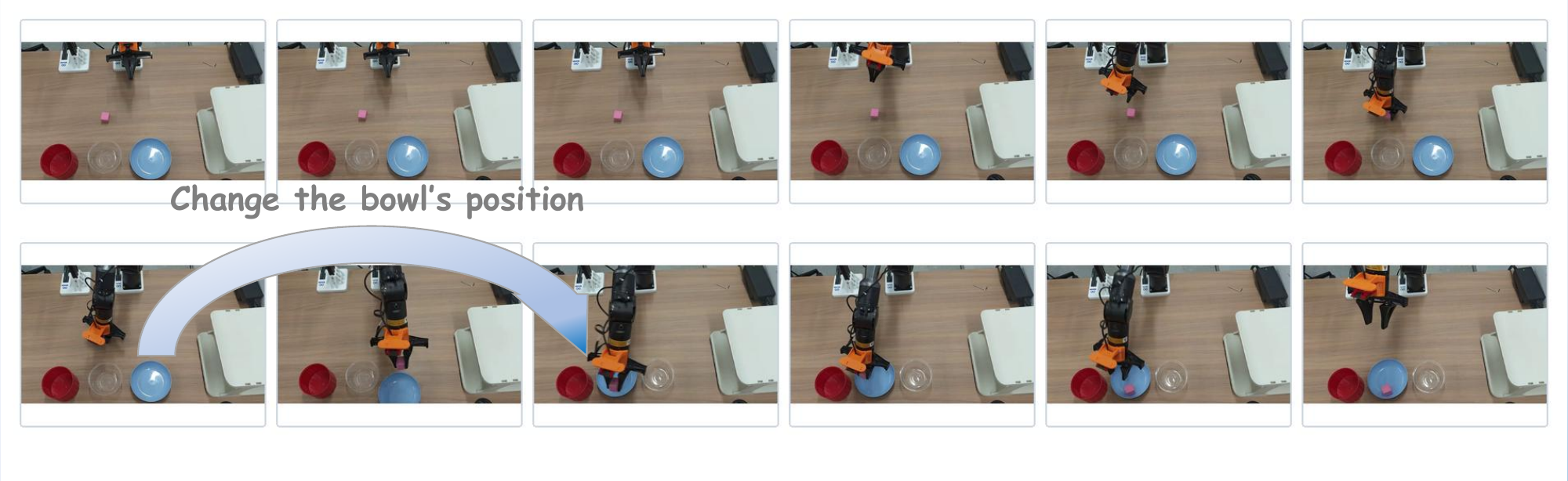}
    \caption{\textbf{Demo: Moving-placement target.} The robot picks up the cube and places it into the blue bowl, while a human manually changes the bowl's position during the process.}
    \label{fig:demo_moving_placement_target}
\end{figure*}

\begin{figure*}[htbp]
    \centering
    \includegraphics[width=0.95\textwidth]{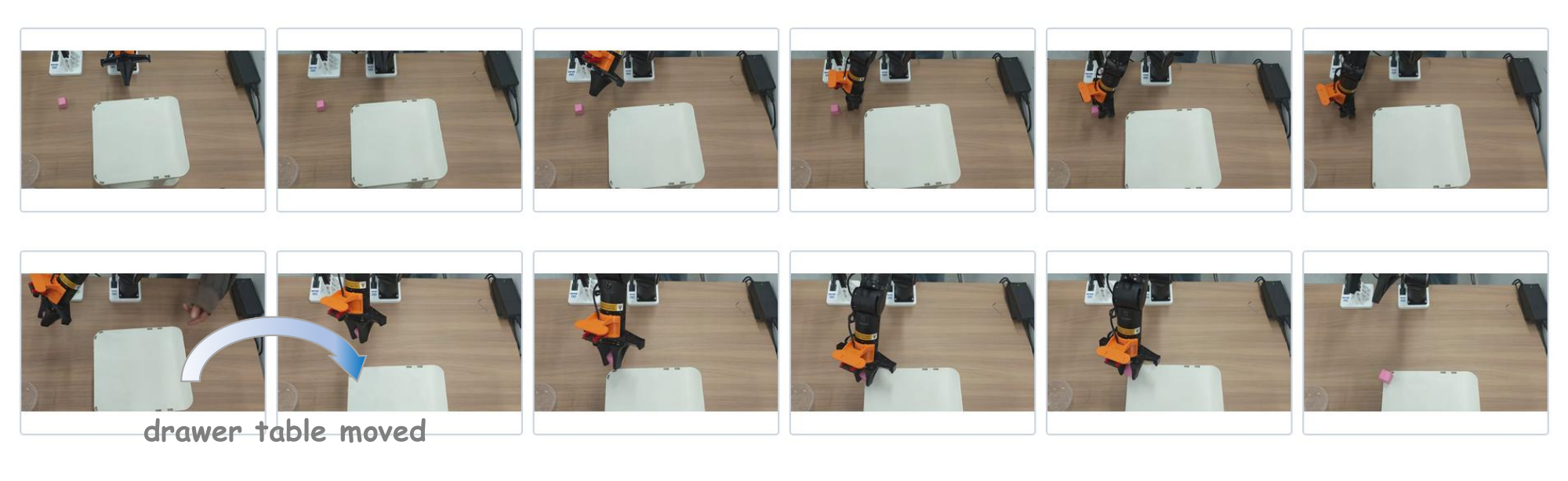}
    \caption{\textbf{Demo: Moving-insertion target.} The robot picks up the cube and places it at the upper-left corner of the drawer top, while a human manually changes the drawer's position during the process.}
    \label{fig:demo_moving_insertion_target}
\end{figure*}

\clearpage

\end{document}